\documentclass[preprint,review,3p]{elsarticle}
\usepackage{natbib}

\newcommand{\cb}[1]{\ifmmode {\boldsymbol{#1}}\else ${\boldsymbol{#1}}$\fi}
\newcommand{\cp}[1]{\ifmmode {\mathcal{#1}}\else ${\mathcal{#1}}$\fi}

\usepackage{booktabs}

\usepackage[T1]{fontenc}
\usepackage{psfrag,epsfig,graphics}
\usepackage{psfrag}
\usepackage{multirow}
\usepackage{multicol}
\usepackage{color}
\usepackage{colortbl}
\usepackage[centertags]{amsmath}
\usepackage{amsfonts}
\usepackage{amssymb}
\usepackage{newlfont}
\usepackage{graphicx}
\usepackage{amsmath}
\usepackage{tabularx}
\usepackage{float}
\usepackage{color, colortbl}
\definecolor{Gray}{gray}{0.9}
\usepackage{comment}
\usepackage{gensymb}
\usepackage{soul}
\usepackage{colortbl}
\definecolor{Gray}{gray}{0.9}
\newcolumntype{a}{>{\columncolor{Gray}}c}

\usepackage[ruled]{algorithm2e}%linesnumbered,
\usepackage{lineno}
\usepackage{svg}
\usepackage{longtable}
\usepackage{lscape}
\usepackage{multirow}
\usepackage{rotating}
\usepackage[colorlinks]{hyperref}
\usepackage[withpage]{acronym}
\usepackage{color,soul}

\usepackage{caption}
\usepackage{subcaption}
\definecolor{Gray}{gray}{0.9}
\soulregister\cite7
\soulregister\ref7
\soulregister\pageref7

\newcolumntype{R}[1]{>{\RaggedRight}p{#1}}

\usepackage{mathrsfs}
 \makeatletter
    \def\ps@pprintTitle{%
       \let\@oddhead\@empty
       \let\@evenhead\@empty
       \let\@oddfoot\@empty
       \let\@evenfoot\@oddfoot
    }
    \makeatother

\usepackage{setspace}
\journal{}

\begin{document}

\begin{frontmatter}

\title{Attention-Guided Fusion of 1D and 2D CNNs for Robust ECG-Based Biometric Recognition}

\author[a]{Islameddine Arioua}
\author[b]{Amir Benzaoui}
\author[c]{Abdelhafid Zeroual}
\author[d]{Lotfi Houam}
%\author[d]{Nabil Hezil}

\address[a]{PIMIS Laboratory, Electronics and Telecommunications Department, Université du 8 Mai 1945, Guelma, Algeria}
\address[b]{Electrical Engineering Department, University of 20 August 1955, Skikda, Algeria} 
\address[c]{Department of Electrical Engineering, Faculty of Science and Applied Sciences, Larbi Ben M'hidi University, Algeria.}
\address[d]{Department of Electronics and Communications, University of Larbi Tebessi, Tebessa, Algeria}

\begin{abstract}
\doublespacing
Electrocardiogram (ECG) biometric recognition has emerged as a promising modality for secure authentication and liveness detection. However, most existing approaches rely on unimodal deep learning architectures that process either one-dimensional (1D) temporal signals or two-dimensional (2D) time-frequency representations independently, limiting their practical robustness and generalizability. To address this limitation, we propose a novel hybrid framework that synergistically integrates 1D and 2D convolutional neural networks (CNNs) within a unified, end-to-end trainable architecture. The 1D CNN pathway extracts temporal dynamics and morphological features from raw ECG waveforms, while the 2D CNN pathway derives discriminative spectral patterns from time-frequency representations. The key contribution of this work is an attention-guided fusion mechanism that enables dynamic, input-dependent weighting of the two modalities, overcoming the limitations of conventional static fusion strategies. The proposed system was rigorously evaluated on three public benchmark datasets—ECG-ID, MIT-BIH, and PTB—encompassing both healthy subjects and patients with cardiac pathologies, achieving state-of-the-art identification accuracies of 99.56\%, 100.00\%, and 99.89\%, respectively. Furthermore, to assess long-term biometric permanence, the framework was evaluated on the multi-session Heartprint dataset collected over ten years, achieving same-session accuracies of 98.54\% (S1), 99.09\% (S2), 94.93\% (S3R), and 96.08\% (S3L), and cross-session accuracies of 56.33\% (S1→S2) and 53.27\% (S2→S3R), demonstrating its ability to capture permanent biometric signatures over extended periods. The optimal configuration employs InceptionTime for 1D processing, ResNet-34 for 2D analysis, and attention-based fusion. Ablation studies confirm that the attention mechanism dynamically recalibrates modality contributions on a per-instance basis, consistently surpassing conventional fusion techniques. Overall, this work establishes a robust, scalable, and end-to-end trainable framework that advances the state-of-the-art in ECG-based biometric recognition.
\end{abstract}

\end{frontmatter}
\section{Introduction}

Biometric recognition systems have increasingly turned to physiological signals to ensure secure and reliable authentication~\cite{amrouni2023palmprint}. Electrocardiogram (ECG)-based methods have gained prominence due to their advantages and promising strategies~\cite{alduwaile2021using, zeroual2025hybrid}. Their benefits include liveness detection, continuous authentication capability, and resilience against spoofing attacks. Unlike conventional biometric modalities such as fingerprints or facial recognition, ECG signals provide dynamic and intrinsic physiological features that are difficult to replicate~\cite{hejazi2016ecg}. Despite these advantages, ECG biometric systems continue to face significant challenges in achieving consistent performance across diverse populations and real-world operational conditions. Current limitations stem from traditional approaches that rely on custom features or unimodal deep learning architectures. Such methods frequently fail to capture the full spatiotemporal complexity of ECG signals. This leads to suboptimal accuracy and limited generalizability ~\cite{li2020toward, carvalho2024addressing}.

\medskip

Recent advances in deep learning have demonstrated the power of Convolutional Neural Networks (CNNs) for ECG signal analysis~\cite{zhang2021human, asadianfam2024ecg}. Current implementations use one-dimensional (1D) CNNs for temporal signal processing or two-dimensional (2D) CNNs for time-frequency analysis. Each of these approaches presents distinct strengths and suffers from specific limitations. While 1D CNNs present an effective modeling of temporal dynamics, they demonstrate high sensitivity to noise artifacts and exhibit limited spectral feature extraction capabilities~\cite{jyotishi2021ecg}. In contrast, 2D CNNs outperform frequency domain analysis but frequently lose critical temporal resolution during signal transformation~\cite{uwaechia2021comprehensive}. Although hybrid approaches combining these paradigms have been explored, most employ static fusion strategies,such as simple feature concatenation or fixed-weight score combination \cite{ullah2021hybrid, kim2025novel, xiaolin2023classification}, that do not account for input-dependent variations in signal quality or discriminative information content.

\medskip

To address these limitations, we propose a novel hybrid architecture that jointly integrates 1D and 2D CNN paradigms within a unified and end-to-end trainable framework for ECG-based biometric recognition. The proposed dual-branch architecture simultaneously processes raw temporal ECG signals and their corresponding time-frequency representations, enabling the extraction of complementary temporal and spectral features. Furthermore, an attention-guided fusion mechanism is introduced to dynamically and adaptively combine these modalities, allowing the model to emphasize the most informative representation on a per-instance basis. This design enhances robustness to signal variability, noise, and heart rate fluctuations, while improving overall recognition performance. The key contribution of this work lies in the synergistic integration of multimodal feature extraction with an adaptive fusion strategy that overcomes the limitations of conventional static fusion approaches.

\medskip

To further enhance overall recognition system performance, first, we explore the impact of ECG segmentation strategies on classification outcomes. Four distinct segmentation techniques are evaluated: R-R interval-based segmentation, R-peak-centered windowing, P-T wave-based segmentation, and fixed-length random segmentation. This comprehensive comparative analysis allows the identification of the most effective preprocessing strategies tailored to the proposed fusion architecture. For a rigorous evaluation of the proposed 1D–2D fusion CNN architectures (feature-level integration, score-level combination, and attention-based fusion), an extensive empirical analysis was conducted against eight state-of-the-art models. This assessment encompasses four advanced 2D CNN architectures, include EfficientNetV2~\cite{tan2021efficientnetv2}, ResNet-34~\cite{koonce2021resnet}, Lightweight CNN~\cite{alduwaile2021using}, and Vision Transformer (ViT)~\cite{azad2024advances}, alongside four sophisticated 1D CNN models, include ResNet-1D~\cite{ganiga2024resnet1d}, InceptionTime~\cite{crocker2021inceptiontime}, ECGNet~\cite{mousavi2019ecgnet}, and XCM~\cite{fauvel2021xcm}. Model validation was conducted using three established benchmark datasets, including ECG-ID~\cite{lugovaya2005biometric}, MIT-BIH~\cite{moody2001impact}, and PTB~\cite{bousseljot1995nutzung}, which collectively represent both controlled laboratory conditions and diverse clinical environments. To rigorously assess the long-term permanence and temporal robustness of the proposed framework, we additionally conduct experiments on the Heartprint dataset %\cite{islam2022heartprint}
, which comprises 1539 ECG recordings collected from 199 healthy subjects over a period of ten years, with intervals of up to 3432 days (approximately 9.4 years) between sessions. This dataset is uniquely suited for evaluating biometric permanence under realistic conditions where enrollment and authentication may occur years apart. The proposed approach addresses three critical gaps in current ECG biometrics systems: (1) the inherent limited representation capacity of unimodal networks to comprehensively capture the divers ECG biometric features; (2) the absence of robust methodologies for managing noisy or variable ECG recordings encountered in practical deployment scenarios; and (3) the critical requirement for systems demonstrating consistent and reliable performance across both healthy and pathological subject populations. 
Accordingly, the principal contributions of this research are fourfold:

\begin{itemize}
  \item \textbf{A novel hybrid 1D-2D CNN framework with attention-guided fusion:} We propose a dual-pathway architecture that synergistically integrates temporal features from raw ECG signals (via a 1D CNN branch) and time-frequency features from scalogram representations (via a 2D CNN branch). Unlike existing static fusion approaches, we introduce an attention-based fusion mechanism that dynamically computes input-dependent modality weights $\alpha$, enabling the model to adaptively prioritize the most informative representation on a per-instance basis (Section 3.4.3). Experimental results (Section 4.5) demonstrate that attention-based fusion consistently outperforms feature-level and score-level fusion, achieving the highest identification accuracy across all three datasets.
  \item \textbf{Comprehensive evaluation of state-of-the-art 1D and 2D architectures:} We systematically evaluate four 1D CNN architectures (InceptionTime, ResNet-1D, ECGNet, XCM) and four 2D CNN architectures (ResNet-34, EfficientNetV2, Lightweight CNN, ViT) under identical experimental conditions. Our analysis identifies InceptionTime and ResNet-34 as the optimal configurations for 1D and 2D processing, respectively, based on performance across multiple segmentation strategies (Sections 4.3 and 4.4).
  \item \textbf{State-of-the-art performance across three benchmark datasets:} The proposed attention-based fusion model achieves identification accuracies of 99.56\% on ECG-ID, 100.00\% on MIT-BIH, and 99.89\% on PTB, surpassing existing methods in the literature (Section 4.7, Table 6).
  \item \textbf{Multi-session evaluation using the Heartprint dataset:} 
    We evaluate the proposed framework on the Heartprint dataset, which contains ECG recordings collected from 199 healthy subjects over ten years (average interval of 1572 days between S1 and S3L). Under same-session protocols, our method achieves 98.54\% (S1), 99.09\% (S2), 94.93\% (S3R), and 96.08\% (S3L) accuracy, outperforming the baseline HeartprintCNN. Under the challenging cross-session protocols (Section 4.8), our method achieves competitive results including 53.27\% for S2→S3R and 51.90\% for S2→S1, demonstrating the framework's ability to capture permanent biometric signatures over extended periods.
    
  \item \textbf{Ablation validation of fusion strategies and robustness analysis:} We conduct a comprehensive ablation study comparing three fusion strategies (feature-level, score-level, and attention-based), demonstrating that the proposed attention mechanism provides the highest accuracy and most stable performance across datasets (Section 4.5). 
\end{itemize}

Experimental results demonstrate that the proposed hybrid methodology significantly outperforms conventional unimodal networks, achieving near-optimal identification accuracy while addressing fundamental limitations of existing ECG biometric systems. This research significantly advances the field of physiological biometrics and provides a generalizable framework for multimodal biosignal processing, with potential applications across diverse biomedical domains.

\medskip

The manuscript organization proceeds as follows: Section \ref{sec2} presents a comprehensive literature review of existing ECG-based biometric methodologies. Section \ref{sec3} details the proposed system architecture, segmentation approaches, and experimental methodology. Section \ref{sec4} discusses experimental results, comparative evaluations, and detailed performance analysis. Section \ref{sec5} concludes the study by summarizing key findings and delineating future research directions.

\section{Related work}\label{sec2}

Current research in ECG-based biometric authentication, often referred to as "heart-print" identification, has increasingly leveraged deep learning methodologies. These advanced techniques can be categorized into one-dimensional (1D) and two-dimensional (2D) approaches, based on their feature extraction mechanisms using CNNs.

\subsection{One-dimensional methods}

1D methods directly process raw temporal ECG signals by one-dimensional convolutional layers to extract waveform morphology and temporal dynamics. Researchers have proposed several advances in this domain. For example, Jyotishi and Dandapat~\cite{jyotishi2021ecg} introduced a comprehensive Long Short-Term Memory (LSTM) architecture for ECG-based person identification. This framework captures both intra-beat and inter-beat variations without requiring fiducial point detection, demonstrating strong performance using short ECG segments. Yuniarti et al.~\cite{yuniarti2024single} proposed a lightweight 1D CNN framework that demonstrated the sufficiency of a single heartbeat segment, interpolated between R-peaks, for robust authentication. Wang et al.~\cite{wang2024ecg} introduced a self-supervised 1D CNN that holds unlabeled data using contrastive learning. The proposed paradigm reduced dependency on annotated datasets while improving generalization.

\medskip

Further advancements include CardioGuard, a hybrid model developed by Ahmed et al.~\cite{ahmed2024cardioguard}, which combines 1D CNNs and LSTMs to extract richer and more discriminative features from ECG signals for classification purposes. In addition, Kim et al.~\cite{kim2023one} presented a fiducial-independent ECG biometric system, employing a 1D neural network to process non-overlapping ECG segments. Their system integrates denoising, segmentation, and classification steps, enabling efficient authentication without the need for fiducial point detection. Melzi et al.~\cite{melzi2023ecg} proposed ECGXtractor, a generalized 1D autoencoder architecture. This model was trained on large-scale datasets to learn transferable representations. Furthermore, Zehir et al.~\cite{zehir2024empirical} developed a hybrid model by the integration of Gated Recurrent Units (GRU) and LSTMs for ECG analysis. Their methodology incorporates bandpass filtering and empirical mode decomposition to extract intrinsic mode functions. This hybrid-recurrent paradigm enables robust multi-scale modeling of both short- and long-term temporal dependencies. It thus addresses the non-stationary and complex nature of ECG signals. Ma et al.~\cite{ma2024out} introduced a dual-branch ECG biometric framework called ORGNNFL (Out-of-distribution Representation and Graph Neural Network Fusion Learning). This approach integrates latent distribution features with topological information derived from 1D ECG signals. In ORGNNFL, an adaptive attention mechanism is used to combine these complementary representational modalities and improve classification accuracy.

\subsection{Two-dimensional methods}

2D methods transform ECG signals into time-frequency representations (such as spectrograms or scalograms) and utilize two-dimensional convolutional architectures to extract spatially organized features. These techniques are particularly effective in capturing spectral patterns and mitigating noise. Ivanciu et al.~\cite{ivanciu2021ecg} implemented a Siamese network architecture for ECG image authentication in distributed environments, demonstrating the feasibility of cloud-based biometric security solutions.  Fuster-Barceló et al.~\cite{fuster2022elektra} proposed the Elektrokardiomatrix (EKM) method, which converts R-peak-aligned ECG segments into heatmaps and matrix structures. These organized inputs were subsequently processed through CNN-based one-against-many identification systems for individual classification. To improve feature extraction efficacy, El Boujnouni et al.~\cite{el2022wavelet} combined Continuous Wavelet Transforms (CWT) techniques with capsule network architectures. Their model achieves high performance by exploiting both local and hierarchical feature relationships. 
Al-Jibreen et al.~\cite{al2024person} designed a lightweight 2D CNN architecture for ECG-based biometrics that accommodates nine distinct arrhythmia classifications. The proposed framework achieves a balance between authentication accuracy and computational efficiency, making it well-suited for practical deployment in clinical environments. Furthermore, Yeşilkaya and Guest~\cite{biccakci2025activity} explored adaptive deep learning techniques incorporating prior activity classification for improved biometric verification performance. Their approach explored multiple time-frequency transformations and hyperparameter tuning to enhance recognition accuracy under dynamic physiological conditions.

\subsection{ Research gap and main contribution}

While 1D CNNs demonstrate effectiveness in processing raw ECG signals through temporal feature extraction, they exhibit limitations. These include insufficient spectral analysis capabilities, vulnerability to waveform morphological variability, and heightened sensitivity to noise artifacts. Conversely, 2D CNNs applied to transformed ECG representations provide superior signal patterns analysis and enhanced noise resilience. Nonetheless, these models often experience loss of precise temporal information, require extensive preprocessing procedures, and demonstrate significant dependence on transformation parameter selection.

\medskip

Although some studies have explored the combination of temporal and transformed ECG representations, most existing approaches rely on static fusion strategies, such as feature concatenation or score-level aggregation, which do not account for the varying importance of each modality across different inputs. This limitation is particularly critical in ECG biometric recognition, where signal quality, noise levels, and morphological characteristics can vary substantially across subjects, recording sessions, and physiological conditions.

\medskip

To address these limitations, this paper proposes a novel hybrid 1D–2D CNN framework that goes beyond conventional multimodal integration by introducing an attention-guided fusion mechanism. This mechanism enables dynamic and input-dependent weighting of temporal and time-frequency features, allowing the model to adaptively prioritize the most informative modality on a per-instance basis. In addition, a feature projection strategy is employed to map heterogeneous features from the 1D and 2D branches into a shared latent space, ensuring effective interaction between modalities and enabling seamless fusion.

\medskip

Furthermore, this work provides a systematic evaluation of multiple ECG segmentation strategies—including P–T segments, QRS-centric windows, R–R intervals, and random fixed-length segments—to identify the most physiologically informative configuration for biometric recognition. A unified comparison of several state-of-the-art 1D architectures (InceptionTime, ResNet-1D, ECGNet, XCM) and 2D architectures (ResNet-34, EfficientNetV2, Lightweight CNN, ViT) is also conducted under consistent experimental conditions.

\medskip

These contributions collectively distinguish the proposed approach from existing studies and enable improved robustness, generalization, and recognition performance across diverse datasets. The proposed framework thus establishes a new paradigm for multimodal ECG biometric recognition that combines the complementary strengths of temporal and spectral feature extraction with adaptive, input-aware fusion.

\section{Proposed approach}\label{sec3}

This section presents the proposed hybrid deep learning architecture, designed for robust ECG-based biometric authentication. This architecture combines temporal and spectral analysis of ECG signals by the integration of two complementary deep learning paradigms, 1D and 2D CNN networks. The proposed architecture uses a 1D CNN for processing raw temporal ECG signals to extract morphological and sequential features. At the same time, a 2D CNN is employed to analyze scalogram representations for comprehensive time-frequency characteristic extraction.

The development of this hybrid model was preceded by a comprehensive evaluation to identify the most effective architectural components. We evaluated the performance of four state-of-the-art 1D CNNs and four 2D CNNs on the task of ECG recognition. Following this rigorous selection process, the best-performing models from each category were integrated using three distinct fusion strategies: feature-level fusion, score-level fusion, and an attention-based fusion mechanism. This multi-strategy fusion approach ensures enhanced model robustness and superior authentication performance across diverse operational conditions. The following subsections provide a detailed exposition of the proposed framework components.

\subsection{Preprocessing}

Signal quality enhancement presents a critical preprocessing stage for robust ECG-based biometric authentication systems. For optimal ECG signal analysis, we implemented a Golay-Savitzky smoothing filter~\cite{bi2024arima}, a sophisticated polynomial-based approach that achieves superior noise suppression while maintaining critical waveform morphology. This technique offers distinct advantages by maintaining the morphological integrity of fiducial points, particularly the P-QRS-T complexes that contain distinctive biometric signatures.
The Savitzky-Golay filter operates through local polynomial approximation, fitting a polynomial of degree $p$ to a sliding window of $2m+1$ samples centered at each discrete point $n$ using least-squares optimization. The filtered signal $y[n]$ is mathematically expressed as:

\begin{equation}\label{eq1}
  y[n]= \sum_{k=0}^{p} a_k^*~n^k
\end{equation}

\noindent where the optimal polynomial coefficients $a_k^*$ are determined by:

\begin{equation}\label{eq2}
{a_k^* }= \arg_{a_k} \min \sum_{i=-m}^{m} \left(x[n+i]- \sum_{k=0}^{p} a_k i^k\right)^2
\end{equation}

\noindent Here, $x[n]$ represents the raw ECG signal, while ${a_k}$ denotes the polynomial coefficient set.

In our implementation, the Savitzky-Golay filter was configured with a window length of $2m+1=11$ samples and a polynomial order of $p=3$. These parameters were selected empirically to balance noise reduction and signal fidelity. The polynomial order of $p=3$ provides sufficient flexibility to preserve the morphological features of the QRS complex while effectively attenuating high-frequency noise.
This preprocessing step effectively attenuates high-frequency noise and baseline wander while preserving critical temporal and amplitude ECG characteristics. These characteristics include the QRS complexes and T-waves, which are indispensable for the accurate extraction of biometric features. Such Savitzky-Golay filter enhancement proves particularly valuable for subsequent 1D and 2D convolutional neural network processing. Temporal precision significantly impacts the 1D CNN's ability to capture morphological features and directly influences the 2D CNN's capacity to generate meaningful time-frequency representations.

\subsection{Segmentation}

\begin{figure}[H]
  \centering
  \includegraphics[width=0.7\linewidth]{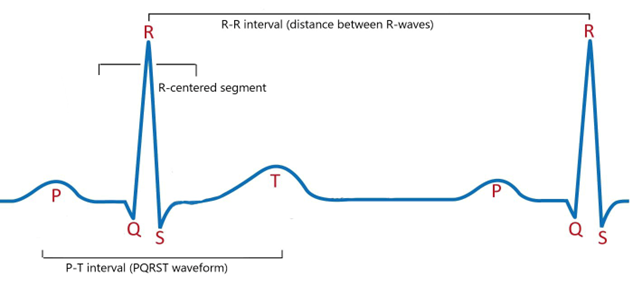}
  \caption{Illustration of the four ECG segmentation strategies: P-T segment, QRS-centric (300 ms), R-R interval, and random fixed-length (1-second) segments.}\label{fig.sig}
\end{figure}

 To systematically evaluate the impact of temporal partitioning on biometric recognition performance, we implemented four clinically-relevant ECG segmentation methodologies. Each targeting distinct physiological characteristics (Figure \ref{fig.sig}):

\begin{itemize}
\item \textbf{P-T Segment}: Segments capturing complete cardiac cycles spanned from P-wave onset to T-wave offset, preserving complete atrial depolarization and ventricular repolarization dynamics.
\item \textbf{QRS-Centric Segment}: Fixed-duration windows of 300 ms centered around detected R-peaks.     
\item \textbf{R-R Intervals}: Two consecutive cardiac cycles capture both beat morphology and rhythm information that reflect individual cardiac conduction patterns.
\item \textbf{Fixed-Duration Control Segments}: 1-second randomly-selected intervals serve as a control condition.
\end{itemize}

For accurate fiducial point detection and precise segmentation boundary determination, we employed a multi-stage approach. First, the Pan-Tompkins algorithm \cite{ihsan2024innovative} was used to achieve robust identification of QRS complexes. The algorithm consists of a sequence of processing steps including band-pass filtering, differentiation, squaring, moving window integration, and adaptive thresholding, enabling reliable R-peak detection even in noisy conditions. It is important to note that the Pan-Tompkins algorithm is specifically designed for QRS detection and does not directly provide the boundaries of P and T waves.

\medskip

Following R-peak detection, the delineation of the P-QRS-T components was performed using a relative temporal windowing strategy based on established physiological ECG characteristics. The QRS complex was defined within a short interval centered around each detected R-peak, with the Q and S points identified as local minima before and after the R-peak within this interval. For P-wave delineation, a search window of 100–200 ms preceding the R-peak was examined. The P-wave onset was defined as the point where the signal first deviates from the isoelectric baseline by more than 10\% of the maximum P-wave amplitude when moving backward from the R-peak, while the P-wave offset was identified as the point following the P-wave peak where the signal returns to baseline. For T-wave delineation, a forward search window of 150–400 ms from the R-peak was used, employing similar slope and amplitude criteria. All detected boundaries were validated against physiologically plausible ranges (P-wave duration: 80–120 ms; T-wave duration: 120–240 ms) to ensure anatomical consistency. These temporal window boundaries were adjusted proportionally to account for variations in heart rate and sampling frequency across datasets. For pathological cases, the windows were visually verified to ensure that the extracted segments captured the relevant waveform components despite morphological variations.

Each segmentation strategy was designed to isolate specific physiological characteristics: the centered R-peak segments targeted QRS complex morphology, P-T segments preserved the complete depolarization-repolarization waveform, and R-R intervals incorporated both morphological and rhythm-based features. The random segments served as a baseline to evaluate the necessity and importance of physiologically-informed segmentation.

All segmented data were subsequently processed using identical feature extraction and classification pipelines, enabling systematic comparison of the segmentation strategies. 

\subsection{Image conversion}

To facilitate robust feature extraction for biometric recognition, 1D ECG signals are converted into 2D time-frequency representations. This study employs the Continuous Wavelet Transform (CWT) to decompose the 1D ECG signal into a joint time-frequency components through convolution with scaled and translated wavelet basis functions:

\begin{equation}\label{eq3}
  W(a,b)=\frac{1}{\sqrt{a}}\int_{-\infty}^{+\infty}x(t)\psi \left(\frac{t-b}{a} \right) dt
\end{equation}

\noindent where $a$ represents the scale parameter (inversely proportional to frequency), $b$ denotes the translation parameter corresponding to temporal shifts, $\psi(.)$ is the complex conjugate of the mother wavelet function and $x(t)$ is the ECG signal. The scale parameter $a$ controls the frequency resolution, with smaller values corresponding to higher frequencies and finer temporal resolution. The relationship between scale $a$ and physical frequency $f$ (in Hz) is given by:

\begin{equation}\label{esca}
  f = \frac{f_c}{a \cdot \Delta t}
\end{equation}

\noindent where, $f_c$ is the center frequency of the mother wavelet (for the Morlet wavelet, $f_c \approx 0.8125$ cycles per sample), $\Delta t$ is the sampling period (inverse of the sampling frequency $f_s$), and $a$ is the scale value. Thus, for a given dataset with sampling frequency \(f_s\), the scale corresponding to a target physical frequency \(f\) is computed as:

\begin{equation}\label{esca1}
  a = \frac{f_c}{f \cdot \Delta t} = \frac{f_c \cdot f_s}{f}
\end{equation}

Rather than directly fixing scale values, we define a physiologically relevant frequency band $[f_{min}, f_{max}]=[0.5,100] Hz$ and discretize it into $N_s=64$ logarithmically spaced frequencies according to:

\begin{equation}\label{logsca}
  f_i = f_{\min} \cdot \left(\frac{f_{\max}}{f_{\min}}\right)^{\frac{i}{N_s-1}}, \quad i = 0, 1, \ldots, N_s-1
\end{equation}

The corresponding scales $a_i$ are then computed using the inverse mapping above, ensuring that the same frequency range is consistently analyzed across all datasets. This formulation guarantees that the CWT representation is adapted to the dataset-specific sampling frequency (ECG-ID: 500 Hz, MIT-BIH: 360 Hz, PTB: 1000 Hz) while preserving identical physiological frequency coverage. Table~\ref{tab:cwt_parameters} presents the dataset-specific parameters.

\begin{table}[H]
\centering
\caption{Dataset-specific CWT parameters}
\label{tab:cwt_parameters}
\begin{tabular}{lcccc}
\hline
\textbf{Dataset} & \textbf{\(f_s\) (Hz)} & \textbf{Nyquist (Hz)} & \textbf{\(a_{\min}\)} & \textbf{\(a_{\max}\)} \\
\hline
ECG-ID & 500 & 250 & 4.06 & 812.50 \\
MIT-BIH & 360 & 180 & 2.93 & 585.00 \\
PTB & 1000 & 500 & 8.13 & 1625.00 \\
Heartprint & 250 & 125 & 2.03 & 406.25 \\
\hline
\end{tabular}
\end{table}

The upper frequency bound is selected such that $f_{\max} \leq f_s/2$ for all datasets, ensuring strict compliance with the Nyquist criterion and preventing aliasing effects. The chosen frequency range captures all relevant ECG components, including low-frequency baseline and $P/T$ wave activity $(\approx 0.5-5 Hz)$, dominant QRS complex energy ($\approx 5-40 Hz$), and higher-frequency morphological details (up to $100 Hz$). The choice of 64 scales provides a frequency resolution of approximately 1.56 Hz, which is sufficient to distinguish between adjacent frequency components of ECG signals. This number of scales also represents an optimal balance between time-frequency resolution and computational efficiency, making the approach practical for processing large-scale datasets. Moreover, the use of 64 logarithmically spaced scales is a standard configuration for CWT-based ECG analysis, consistent with established practices in the literature \cite{islam2022heartprint}.

Unlike the Discrete Wavelet Transform (DWT), which provides coefficients at discrete dyadic scales, the CWT produces a continuous time-frequency map by evaluating the wavelet at all scales, enabling fine-grained spectral analysis. The CWT configuration was adapted to the sampling frequency of each dataset to ensure consistent time-frequency representation. The following parameters were used:
      
      \begin{itemize}
        \item Mother wavelet: Morlet wavelet, chosen for its optimal time-frequency localization properties and its effectiveness in capturing oscillatory patterns characteristic of ECG signals.
        \item Scales: 64 logarithmically spaced scales covering the physiologically relevant ECG frequency range (0.5–100 Hz).
        \item Scalogram size: 224×224 pixels, standardized to match the input dimensions required by pre-trained 2D CNN architectures.
      \end{itemize}
      
\begin{figure}[H]
     \centering
     \begin{subfigure}[b]{0.49\textwidth}
         \centering
         \includegraphics[width=0.8\textwidth]{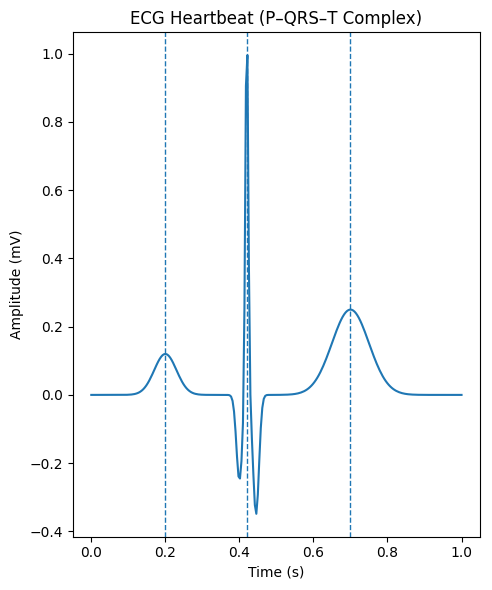}
         \caption{A single heartbeat segment (1D ECG signal)}
         \label{hb}
     \end{subfigure}
     \hfill
     \begin{subfigure}[b]{0.50\textwidth}
         \centering
         \includegraphics[width=0.8\textwidth]{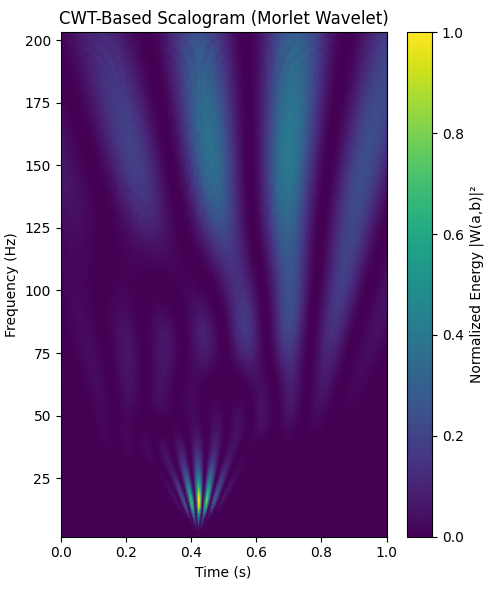}
         \caption{Resulting scalogram from CWT (2D time-frequency representation)}
         \label{trcwt}
     \end{subfigure}
       \caption{Illustration of the transformation of a 1D ECG signal into a 2D time-frequency representation (scalogram) using the Continuous Wavelet Transform (CWT). The scalogram reveals time-frequency patterns where P-QRS-T complexes manifest as distinctive high-energy coefficient regions.}
        \label{fig:cwt}
\end{figure}

From the CWT coefficients, the squared magnitude \(|W(a,b)|^2\)  was computed to generate a 2D time-frequency scalogram representation, which was subsequently normalized to a standard range. This transformation enables the identification of discriminative spectral signatures within QRS complexes, P-waves, and T-waves that remain consistent across different recording conditions while exhibiting substantial inter-individual variability.  
Figure \ref{fig:cwt} illustrates the transformation of a single heartbeat from its raw 1D ECG signal to its corresponding 2D time-frequency scalogram using the CWT. The scalogram reveals distinct patterns corresponding to the P-QRS-T complexes, which serve as discriminative features for biometric recognition.

CWT allows 2D CNNs to capture complex spatial patterns in the spectral domain and preserves temporal localization. These features complement the temporal patterns extracted by the 1D CNN pathway in the proposed dual-pathway framework.
 	 
\subsection{Feature extraction}

Feature extraction constitutes the core computational component of the proposed approach. It centered on a comparative evaluation of CNNs operating on two distinct representations of the ECG signal. First, to capture temporal features, four distinct 1D CNN architectures were trained and evaluated directly on the raw ECG waveforms. Alongside, four advanced 2D CNN models were applied to analyze scalogram-based time-frequency representations.A comprehensive evaluation methodology involves individual performance assessment of each architectural configuration to identify optimal feature extraction. Subsequently, the best-performing models from both 1D and 2D CNN categories are strategically selected for integration through the fusion mechanisms in the subsequent processing stage. The specific architectures and configurations of all eight evaluated CNN models are detailed in the subsequent subsections.

\subsubsection{1D CNNs}\label{sub3.4.1}

To extract features directly from the ECG raw signal, four prominent 1D CNN architectures were employed. These models were selected for their proven efficacy in processing sequential data and their ability to learn hierarchical temporal patterns.

\subsubsection*{ResNet-1D}

An adaptation of the residual network architecture for 1D temporal signal processing \cite{ganiga2024resnet1d}. ResNet-1D utilizes residual connections that facilitate gradient flow through deep networks. This architecture was chosen for its capacity to mitigate the vanishing gradient problem. ResNet-1D allows the training of deeper architectures while maintaining computational stability and convergence properties.

\subsubsection*{InceptionTime}

InceptionTime \cite{crocker2021inceptiontime} is a specialized architecture for time series classification inspired by the Inception module from computer vision. InceptionTime employs multiple parallel convolutional filters with varying kernel sizes to capture temporal features across diverse time scales simultaneously. Its inclusion is justified by the natural variability in the duration of cardiac events (e.g., QRS complexes, T-waves) across different individuals, which requires multi-resolution feature extraction.

\subsubsection*{ECGNet}

ECGNet \cite{mousavi2019ecgnet} is a deep learning model specifically designed for ECG signal classification and analysis. Their architecture integrates convolutional layers with attention mechanisms or recurrent layers to automatically identify important features within the ECG signal. ECGNet demonstrated its ability to effectively detect and classify critical cardiac signatures, making it highly suitable for discerning person-specific biometric markers.

\subsubsection*{XCM}

The Explainable Convolutional Network (XCM) for multivariate time series classification~\cite{fauvel2021xcm} is an interpretable deep learning framework that combines multivariate time series analysis with explainability mechanisms. XCM is designed to learn both temporal patterns and the relative importance of different features. The architecture utilizes 1D convolutions integrated with channel-wise attention mechanisms to learn both temporal patterns and feature importance simultaneously. The channel-wise attention component enables dynamic weighting of different ECG lead contributions, facilitating adaptive feature selection based on signal quality and discriminative power.

\subsubsection{2D CNNs}\label{sub3.4.2}

To extract features from the two-dimensional scalogram representations, four leading image classification architectures were utilized. These models were selected for their state-of-the-art performance in visual pattern recognition.

\subsubsection*{EfficientNetV2}

EfficientNetV2 \cite{tan2021efficientnetv2} is a state-of-the-art CNN that achieves high accuracy with superior parameter efficiency. Their architecture surpasses the original EfficientNet through the incorporation of progressive learning strategies, enhanced compound scaling, and fused mobile inverted bottleneck convolutions. These architectural innovations significantly improve training efficiency and accuracy while reducing computational overhead. EfficientNetV2 achieves high accuracy with fewer parameters, making it efficient for ECG scalogram analysis.

\subsubsection*{ResNet-34}

ResNet-34 \cite{koonce2021resnet} is a 34-layer CNN variant of the Residual Network family, widely recognized for its strong balance between network depth, accuracy, and computational load. ResNet-34 architecture employs residual connections to facilitate training of deep architectures while avoiding the vanishing gradient issues. It was included as a robust and well-established baseline for image classification tasks.

\subsubsection*{Lightweight CNN}

Lightweight CNN \cite{alduwaile2021using} refers to a class of CNNs optimized for minimal computational cost and a small memory footprint. This property makes them ideal for deployment on mobile or embedded devices. These models typically employ techniques like depthwise separable convolutions, global average pooling, and parameter reduction techniques to minimize computational overhead while maintaining feature extraction efficacy.

\subsubsection*{ViT}

Vision Transformer (ViT) \cite{azad2024advances} is a model that applies the transformer architecture, originally from natural language processing, to image analysis. ViT partitions input images into fixed-size patches, treats them as sequential tokens, and processes them through self-attention mechanisms. This approach enables global context modeling and long-range dependency capture that surpasses traditional convolutional architectures. For ECG scalogram analysis, ViT's self-attention mechanisms facilitate comprehensive analysis of time-frequency relationships across the entire cardiac cycle.

\subsubsection{Multi-modal fusion strategies}

The integration of complementary features extracted from temporal and spectral ECG representations constitutes the fundamental innovation of the proposed dual-pathway architecture. This section presents three proposed fusion methodologies that synergistically combine discriminative characteristics from both 1D and 2D CNN processing pathways: feature-level integration, score-level combination, and attention-mechanism-based fusion. Each fusion strategy employs distinct computational approaches to merge information from raw temporal ECG signals and their corresponding time-frequency representations. This integration aims to improve the system's overall robustness and discriminative capability, thereby ensuring optimal exploitation of complementary information while mitigating the individual limitations of unimodal processing approaches.

\subsubsection*{Feature-level fusion}

Feature-level fusion implements early integration  through the combination of intermediate feature representations extracted from the 1D and 2D CNN processing branches prior to final classification. 
Let $F_1 \in \mathbb{R}^{d_1}$ and $F_2 \in \mathbb{R}^{d_2}$ represent the feature vectors from the 1D and 2D branches, respectively. These vectors are concatenated to form a unified feature vector, $F_{\text{fused}}$:

\begin{equation}\label{eqflf}
F_{fused}=F_1 \oplus F_2 \in \mathbb{R}^{d_1+d_2}
\end{equation}

\noindent where $\oplus$ denotes the concatenation operation. This joint vector is subsequently fed into one or more fully connected layers for classification. 

\begin{figure}[H]
  \centering
  \includegraphics[width=0.45\linewidth]{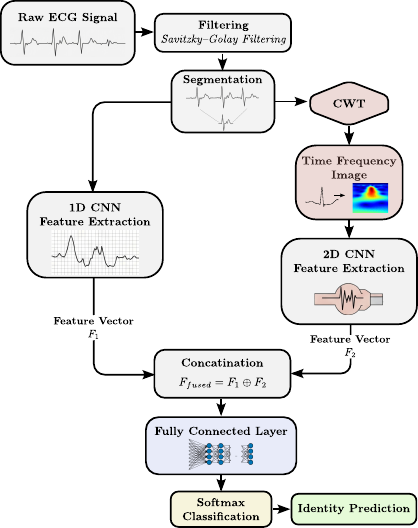}
\caption{Architectural overview of feature-level fusion methodology for ECG biometric authentication. The 1D CNN branch extracts temporal features from raw ECG signals, while the 2D CNN branch extracts spectral features from CWT-based scalograms. The extracted feature vectors are concatenated prior to the classification layer.}\label{flow1}
\end{figure}

While this method preserves the maximum amount of information from both modalities, it forms a high-dimensional feature space and implicitly assumes that the contributions of each modality are static across all samples. Figure \ref{flow1} illustrates this process. The framework demonstrates the integration of temporal features from 1D CNN (processing filtered raw ECG signals) with spectral features from 2D CNN (processing CWT-based time-frequency images) through direct concatenation prior to classification.

\vspace{-0.25cm}

\vspace{-0.75cm}
\subsubsection*{Score-level fusion}

Score-level fusion implements late integration and operates at the decision level by combining the output probability scores from the independently trained 1D and 2D classifiers.

 Let $s_1 \in \mathbb{R}^C$ and $s_2 \in \mathbb{R}^C$ be the softmax-normalized class probability vectors for the $C$ identity classes, generated by the 1D and 2D models, respectively. The final score vector, $s_{\text{fused}}$, is calculated as a weighted sum:

\begin{equation}\label{eqflf}
s_{fused}= \lambda s_1+(1-\lambda)s_2
\end{equation}

\noindent where the hyperparameter $\lambda \in [0,1]$ balances the influence of each modality. The final prediction corresponds to the class with the highest fused score. This approach offers modularity, as each classifier can be trained and optimized independently. However, it may neglect complex inter-modal dependencies that are only present at the feature level. 

\begin{figure}[H]
  \centering
  \includegraphics[width=0.45\linewidth]{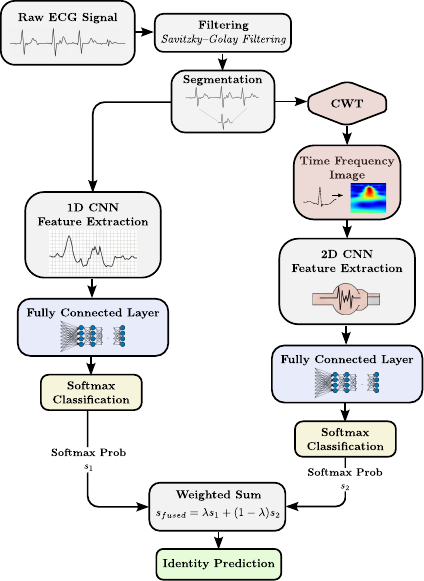}
  \caption{Architectural overview of score-level fusion methodology for ECG biometric authentication. Softmax probability scores \(s_1\) and \(s_2\) from independently trained 1D and 2D CNN classifiers are combined via weighted aggregation \(s_{\text{fused}} = \lambda s_1 + (1-\lambda)s_2\), where the fusion weight \(\lambda\) is optimized via parameter sweep.}\label{flow2}
\end{figure}

Score-level fusion architecture, depicted on Figure \ref{flow2}, demonstrates the combination of prediction probabilities from independently trained CNN classifiers. The framework shows how decision scores from the 1D CNN processing raw ECG signals and the 2D CNN analyzing CWT representations are integrated through weighted aggregation for final identity prediction.

\vspace{-0.75cm}

\subsubsection*{Attention-based fusion}

To enable a more dynamic integration of the two modalities, an attention-based feature fusion mechanism was implemented. This framework applies adaptive weighting to combine complementary temporal and spectral information extracted from ECG signals. This allows the model to adjust the contribution of 1D and 2D features for each input, focusing on the most informative modality. The attention-based fusion framework addresses the fundamental limitation of static fusion approaches by introducing dynamic modality weighting that adapts to each input. This mechanism proves particularly valuable for ECG biometric authentication, where signal quality, recording conditions, and individual physiological characteristics may vary significantly across subjects and scenarios.

Given that the feature vectors $F_1 \in \mathbb{R}^{d_1}$ and $F_2 \in \mathbb{R}^{d_2}$ may have different dimensions, they are first projected into a shared latent space of dimension $d$ using learnable linear transformations:
     
\begin{eqnarray}
 \hat{F}_1 &=& W_1 F_1+b_1 \\
 \hat{F}_2 &=& W_2 F_2+b_2
 \end{eqnarray}
 
\noindent where $W_1 \in \mathbb{R}^{d\times d_1}$ and $W_2 \in \mathbb{R}^{d\times d_2}$ are projection matrices, and $b_1, b_2 \in \mathbb{R}^d$ are bias vectors. This projection ensures dimensional compatibility for the subsequent fusion step. In our implementation, the feature dimensions were $d_1=512$ for the InceptionTime 1D branch and $d_2=512$ for the ResNet-34 2D branch, with a shared latent dimension of $d=256$. A soft attention mechanism then calculates a scalar attention weight, $\alpha$, which quantifies the relative importance of the modalities. The concatenated projected features are passed a two-layer neural network to produce this weight. Specifically, the first layer applies a hyperbolic tangent activation to capture non-linear interactions between modalities:

\begin{equation}
h = \tanh(W_{attn} \cdot (\hat{F}_1 \oplus \hat{F}_2) + b_{attn})
\label{eq:attention_hidden}
\end{equation}

where $W_{attn} \in \mathbb{R}^{d_{attn} \times 2d}$ and $b_{attn} \in \mathbb{R}^{d_{attn}}$ are learnable parameters, with $d_{attn}=128$ representing the hidden dimension. The second layer compresses this representation to a single scalar, followed by a sigmoid activation:

\begin{equation}
\alpha = \sigma(w_{attn}^T h + c)
\label{eq:attention_weight}
\end{equation}

In this formulation, $w_{attn} \in \mathbb{R}^{d_{attn}}$ is a learnable weight vector, $c \in \mathbb{R}$ is a scalar bias, and $\sigma(\cdot)$ is the sigmoid function, which constrains $\alpha$ to the range $[0, 1]$.

In practice, the attention weight $\alpha$ acts as a gating coefficient that dynamically regulates the contribution of the two feature branches. Since the sigmoid activation function constrains $\alpha$ to the range $[0,1]$, the fusion formulation naturally produces a convex combination of the projected feature vectors $\hat{F}_1$ and $\hat{F}_2$. Consequently, larger values of $\alpha$ increase the contribution of the temporal representation extracted by the 1D CNN, whereas smaller values emphasize the spectral representation obtained from the 2D CNN branch.

The final fused feature vector, $F_{\text{fused}}$, is computed as a convex combination of the projected features, moderated by the attention weight:

\begin{equation}
    F_{\text{fused}} = \alpha \hat{F}_1 + (1-\alpha)\hat{F}_2
    \label{eq:attention_fusion}
\end{equation}

This adaptive formulation permits the model to dynamically prioritize the more discriminative modality on an instance-by-instance basis, which can improve robustness against variations in signal quality. Furthermore, the attention weight $\alpha$ enhances model interpretability by providing insight into which modality contributed more to a given prediction. The complete set of network parameters, including the projection and attention layers, is trained end-to-end. 

During training, the parameters of the projection layers ($W_1, W_2, b_1, b_2$) and the attention module ($W_{attn}, b_{attn}, w_{attn}, c$) are jointly optimized with the entire network using backpropagation. As a result, the model automatically learns how to balance temporal and spectral ECG representations depending on the discriminative information contained in each modality. Furthermore, the attention weight $\alpha$ enhances model interpretability by providing insight into which modality contributed more to a given prediction. For instance, in our experiments, we observed that segments with high noise levels in the temporal domain resulted in lower $\alpha$ values (favoring the 2D branch), while segments with clear morphological features produced more balanced weights.

\begin{figure}[H]
  \centering
  \includegraphics[width=0.45\linewidth]{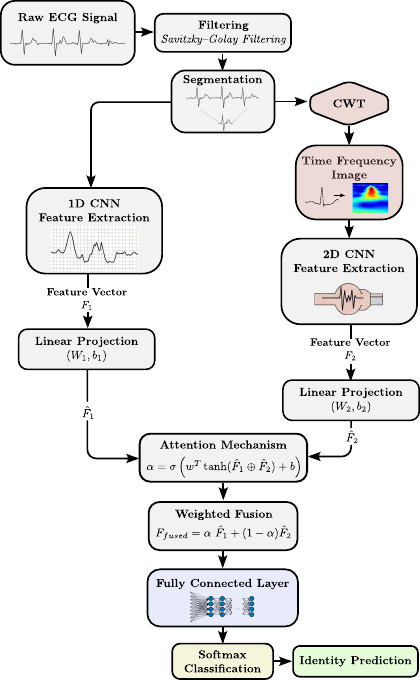}
  \caption{Architectural overview of attention-based fusion methodology for ECG biometric authentication. Projected features \(\hat{F}_1\) and \(\hat{F}_2\) from the 1D and 2D CNN branches are passed through an attention network that computes a dynamic weight \(\alpha\). The final fused feature vector is computed as \(F_{\text{fused}} = \alpha \hat{F}_1 + (1-\alpha)\hat{F}_2\), enabling input-dependent modality weighting.}\label{flow3}
\end{figure}

As illustrated in Figure \ref{flow3}, an attention-based fusion architecture employing soft attention mechanisms for adaptive integration of multi-modal ECG features. The framework demonstrates dynamic weighting of transformed feature vectors from 1D and 2D CNN streams, enabling the model to selectively prioritize the most informative modality for each authentication instance.

In summary, each fusion strategy presents distinct trade-offs between computational complexity, implementation simplicity, and authentication performance. Feature-level fusion provides comprehensive representational integration through early combination; score-level fusion allows for modular classifier design with late-stage decision integration; and attention-based fusion introduces adaptive learning capabilities for dynamic modality importance assessment. A comparative evaluation of these methods is presented in a subsequent section to identify the optimal configuration for ECG-based biometric recognition.

\section{Experimental analysis}\label{sec4}

This section details the empirical evaluation of the proposed framework. It begins by describing the datasets and the computational environment, followed by the experimental protocols and evaluation metrics. Subsequently, it reports the results of the unimodal 1D and 2D CNN models, analyzes the performance of different fusion strategies through an ablation study, and concludes with a comparative assessment against state-of-the-art ECG-based biometric methods.

\subsection{Data description} %and computing platform}

The framework's performance was evaluated and validated using three established ECG benchmark databases: ECG-ID \cite{lugovaya2005biometric}, MIT-BIH Arrhythmia Database \cite{moody2001impact}, and PTB Diagnostic ECG Database \cite{bousseljot1995nutzung}. These datasets collectively represent diverse recording conditions, subject populations, and clinical scenarios, providing comprehensive evaluation coverage for biometric authentication performance.

\medskip

\begin{description}
  \item[ECG-ID Database \cite{lugovaya2005biometric}] This database comprises 310 single-lead ECG recordings acquired from 90 volunteers aged 13-75 years, with balanced gender representation (44 males, 46 females). Each recording spans 20 seconds duration, captured at 500 Hz sampling rate with 12-bit resolution and 10 $mV$ dynamic range. The recordings were captured non-contemporaneously, with intervals up to six months, introducing significant intra-subject variability that presents a robust challenge for biometric systems.
  \item[MIT-BIH Arrhythmia Database \cite{moody2001impact}] This dataset contains 48 dual-channel ambulatory ECG recordings from 47 subjects (25 males, 22 females), with one participant contributing two separate sessions (labeled 201 and 202). Recordings were acquired at 360 Hz sampling frequency with 11-bit resolution and amplitude ranges exceeding 10 $mV$. While this dataset is primarily a benchmark for arrhythmia detection, its physiological diversity is suitable for testing the generalization capabilities of biometric models.
  \item[PTB Diagnostic ECG Database \cite{bousseljot1995nutzung}] This clinical dataset encompasses 549 recordings from 290 subjects, including 52 healthy controls and 238 patients presenting diverse cardiovascular pathologies including myocardial infarction and other cardiac disorders. Data were acquired using a 12-lead setup at 1 kHz with 16-bit resolution. For methodological consistency, only Lead II signals were used in this study.
\end{description}

In this study, Lead II from the PTB Diagnostic ECG Database was selected because it provides a clear and stable representation of the P–QRS–T morphology, which is particularly suitable for reliable heartbeat segmentation and feature extraction. Moreover, Lead II typically exhibits prominent R-peaks, facilitating accurate detection using the Pan–Tompkins algorithm. To ensure consistency across datasets, we followed the lead-selection strategy commonly adopted in related studies. For the MIT-BIH Arrhythmia Database, the Modified Lead II (MLII) signal was used, as it is directly comparable to standard Lead II and is the most commonly utilized lead in studies based on this dataset. For the ECG-ID Database, the available lead corresponds to a standard limb-lead configuration that produces a waveform morphology closely resembling that of Lead II, as confirmed by visual inspection and consistent with previous studies using this dataset. This approach helps maintain comparable ECG morphological characteristics across datasets, thereby ensuring consistent feature extraction and fair evaluation of the proposed method.

\subsection{Experimental protocol and evaluation} %metrics}

The experimental design adheres to evaluation protocols established in recent literature while implementing dataset-specific procedures to ensure methodological consistency, prevent data leakage, and guarantee reproducibility.

\subsubsection{Dataset Preparation}

From the ECG-ID database, the first two recordings per subject were used to ensure consistency across subjects while maintaining the challenge of cross-session variability, as these recordings were acquired non-contemporaneously with intervals up to six months. For the PTB database, the initial 60 seconds of Lead II from the first recording were selected to provide a standardized input length while preserving diagnostic information. For the MIT-BIH database, 60-second segments were extracted from the beginning of each record to maintain uniformity across all datasets. The 60-second segmentation stage (for PTB and MIT-BIH) serves as a preprocessing scaffold to ensure local signal stationarity, standardized preprocessing across heterogeneous datasets, temporally distributed heartbeat extraction, and quality control. Following this intermediate segmentation, each 60-second segment undergoes R-peak detection using the Pan–Tompkins algorithm \cite{ihsan2024innovative}, followed by the extraction of individual heartbeat-level segments (P-T segments, QRS-centric windows, etc.). The final biometric representation remains strictly heartbeat-based.

The selection of 60-second segments for the PTB and MIT-BIH databases follows commonly adopted protocols in ECG biometric studies, ensuring consistency with related work. For the ECG-ID database, the full recordings (approximately 20 seconds) were used due to their fixed duration. It is important to note that all signals are further segmented into individual cardiac cycles (e.g., P–T segments), and each segment is treated as an independent instance. Therefore, the effective inputs correspond to heartbeat-level segments, which reduces the influence of varying recording durations across datasets and ensures that the number of instances per subject remains comparable across datasets.

Following the segmentation methodology outlined in Section 3.2, the extracted temporal segments generated multiple instances per subject for comprehensive evaluation. Table \ref{tab:dataset_summary1} summarizes dataset characteristics including subject counts, total instances, and average instances per subject.

\begin{table}[H]
\centering
\caption{Dataset characteristics summary, showing the number of subjects, total instances, and average instances per subject.}
\label{tab:dataset_summary1}
\begin{tabular}{lccc}
\hline
\textbf{Dataset} & \textbf{\# Subjects} & \textbf{\# Instances} & \textbf{Avg. \# Instances / Subject} \\ \hline
ECG-ID & 90 & 1797 & $\approx$20 \\
MIT-BIH & 48 & 960 & 20 \\
PTB & 290 & 5800 & 20 \\ \hline
\end{tabular}
\end{table}

\subsubsection{Cross-Validation Protocol and Data Leakage Prevention}\label{cross}

A 5-fold cross-validation protocol was employed for all experiments to ensure a robust and unbiased evaluation. To prevent any potential data leakage, the partitioning was performed at the subject level, meaning that all segments extracted from the same subject were assigned exclusively to the same fold. This is a critical design choice because biometric systems must be evaluated on their ability to recognize unseen segments (i.e., instances), not merely on distinguishing between segments from the same subject that the model has already seen during training.

Specifically, for each fold, the instances were randomly partitioned into five approximately equal subsets. Four subsets (80\% of instances) were used for training, while the remaining subset (20\% of instances) was used for testing. This approach guarantees that no instance appears in both the training and testing sets within a given fold, ensuring that the model is evaluated on previously unseen instances. This evaluation strategy provides a realistic assessment of the model's discriminative capability under closed-set conditions while preventing trivial data leakage.

For hyperparameter tuning and model selection, a validation subset was extracted from the training data during each fold. This validation set represents a portion of the training folds and remains fully independent from the test data, ensuring that the test set is used only for final evaluation. All reported results are the mean and standard deviation across the five folds, providing a comprehensive and unbiased assessment of model performance while enhancing the reliability and statistical significance of the reported results.

\subsubsection{Evaluation Metrics}

In this study, the proposed framework is evaluated under the closed-set identification paradigm, which is widely adopted in recent ECG-based biometric recognition literature \cite{el2022wavelet, kim2023one, wang2024ecg, carvalho2024addressing}. In biometric systems, two evaluation paradigms are commonly considered: identification mode, where the system assigns an input to one of several enrolled identities (a multi-class classification problem), and verification mode, where the system validates a claimed identity based on similarity scores (a binary decision problem). Given that our study specifically targets subject recognition in a closed-set setting, the identification protocol was considered the most appropriate. This approach aligns with the dominant practices in recent ECG biometric literature, where identification accuracy serves as the primary performance metric for evaluating discriminative capabilities.

Identification accuracy is used as the primary metric to assess the discriminative capability of the learned representations across a predefined set of subjects. In addition, complementary metrics including precision, recall, and F1-score are reported to provide a more comprehensive evaluation of the model performance across different segmentation strategies and architectural configurations. These metrics are defined as follows:

\begin{equation}\label{acc}
\text{Accuracy} =\frac{TP+TN}{TP+TN+FP+FN}\times 100
\end{equation}

\begin{equation}\label{prec}
\text{Precision}=\frac{TP}{TP+FP}\times 100
\end{equation}

\begin{equation}\label{rec}
\text{Recall}=\frac{TP}{TP+FN}\times 100
\end{equation}

\begin{equation}\label{f1s}
\text{F1-score} =2\times\frac{\text{Precision}\times\text{Recall}}{\text{Precision}+\text{Recall}}\times 100
\end{equation}

\noindent where TP, TN, FP, and FN represent true positives, true negatives, false positives, and false negatives, respectively. For multi-class classification, these metrics are computed per class and macro-averaged to provide a balanced assessment across all subjects.

\subsubsection{Computing Platform}

All experiments were implemented using PyTorch 2.6.0 framework with CUDA 12.4 acceleration, executed within an Anaconda environment. The experimental platform comprised an Intel Core i7-10750H processor, 32 GB RAM, and NVIDIA RTX 4090 GPU (6 GB VRAM), ensuring adequate computational resources for deep learning model training and evaluation.

\subsubsection{Training Configuration and Hyperparameters}

To ensure reproducibility and enable fair comparison across all models, a consistent training protocol was established. All models, including the four 1D CNN architectures (InceptionTime, ResNet-1D, ECGNet, XCM) and the four 2D CNN architectures (ResNet-34, EfficientNetV2, Lightweight CNN, ViT), were trained under identical conditions using the same optimization settings, data splits, and evaluation protocol. The training configuration is summarized in Table \ref{tab:hyperparameters}.

\begin{table}[H]
\centering
\caption{Training hyperparameters and configuration settings}
\label{tab:hyperparameters}
\begin{tabular}{lc}
\hline
\textbf{Parameter} & \textbf{Value} \\
\hline
\textbf{Optimizer} & Adam \\
\textbf{Learning rate} & $1 \times 10^{-3}$ \\
\textbf{Batch size} & 16 \\
\textbf{Maximum epochs} & 50 \\
\textbf{Feature embedding dimension} & 256 \\
\textbf{Dropout rate} & 0.5 (for fully connected layers) \\
\hline
\end{tabular}
\end{table}

To ensure fair comparison, all baseline models were trained using the identical protocol described above. The same cross-validation splits (described in Section \ref{cross}) were used for all models, and performance metrics were computed on the same test sets. This rigorous consistency eliminates variations in training conditions as a confounding factor when comparing architectural performance.

\subsection{1D CNNs results}

An initial comprehensive evaluation was conducted to assess the biometric recognition performance of four state-of-the-art 1D CNNs, including ResNet-1D, InceptionTime, ECGNet, and XCM, across multiple ECG segmentation strategies. As detailed in Section  \ref{sub3.4.1}, each architecture was fine-tuned and evaluated using the following segmentation techniques: (1) P–T segments encompassing a full cardiac depolarization-repolarization cycle; (2) R-centered segments focused on the QRS complex and adjacent S–T intervals; (3) R–R intervals covering two consecutive heartbeats; and (4) random fixed-length segments serving as a physiologically uninformative baseline.

\medskip

The models were rigorously tested on three publicly available ECG datasets including ECG-ID, MIT-BIH, and PTB, under a consistent evaluation protocol incorporating multiple performance metrics (Accuracy, Precision, Recall and F1-score). Figure \ref{1dcomp} presents the results via multi-axis radar charts, enabling a holistic visual comparison of each model’s effectiveness under different segmentation strategies. Note that each radar chart uses adaptive scaling to the performance range of the specific dataset and model configuration to maximize discriminative visibility of high-performance models. Numerical values are provided in Table \ref{tab:1dmean} for precise quantitative comparison.

\begin{figure}[H]
\centering
\subfloat[ECG-ID Dataset]{\includegraphics[width=0.31\textwidth]{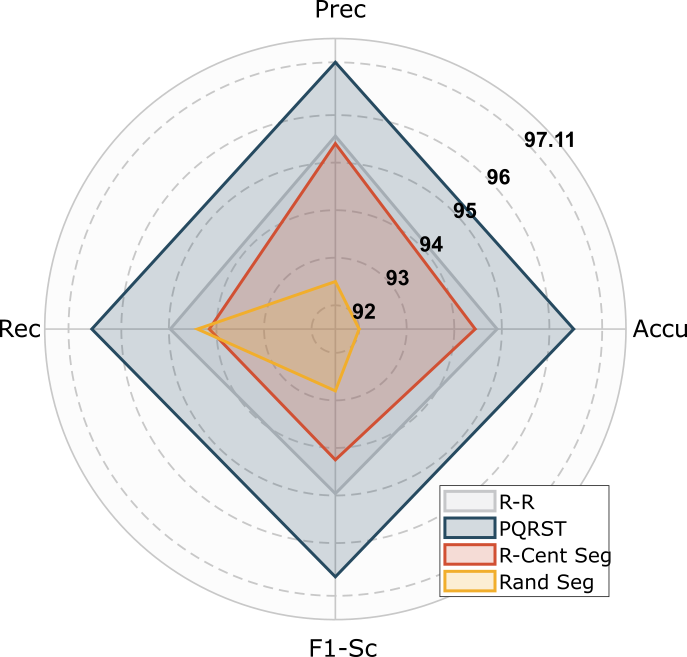}}\hfill
\subfloat[MIT-BIH Dataset]{\includegraphics[width=0.31\textwidth]{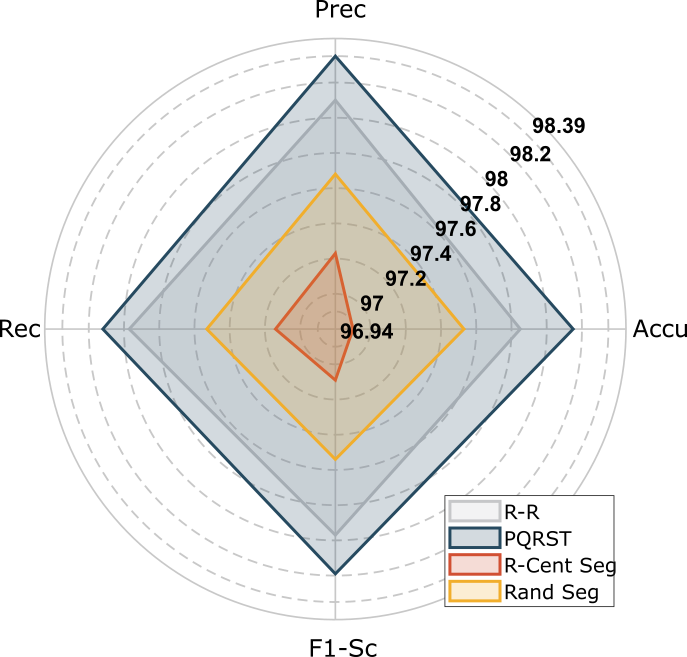}}\hfill
\subfloat[PTB Dataset]{\includegraphics[width=0.31\textwidth]{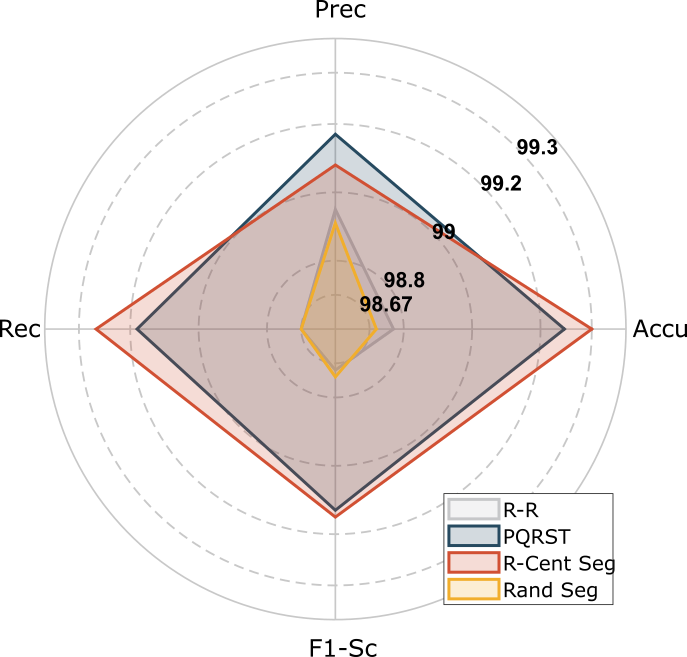}}\\
\caption*{(A) ResNet-1D CNN Architecture}
\end{figure}

\begin{figure}[H]\ContinuedFloat
\subfloat[ECG-ID Dataset]{\includegraphics[width=0.31\textwidth]{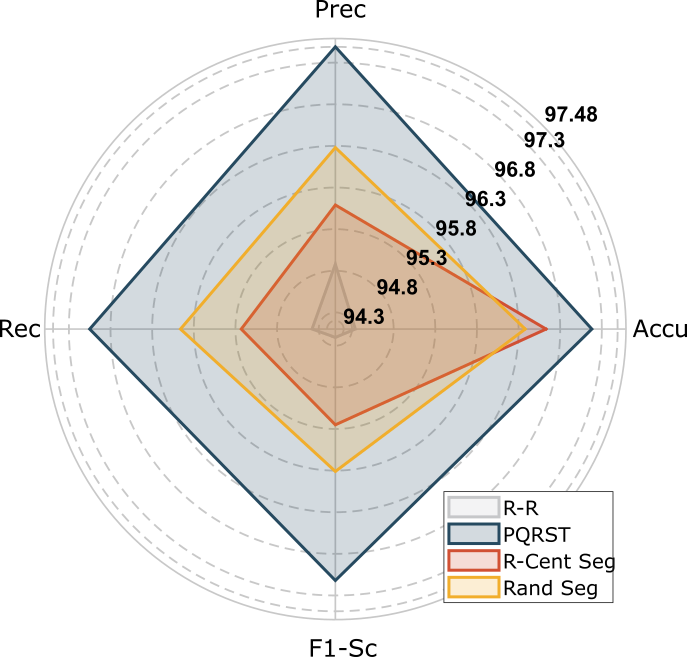}}\hfill
\subfloat[MIT-BIH Dataset]{\includegraphics[width=0.31\textwidth]{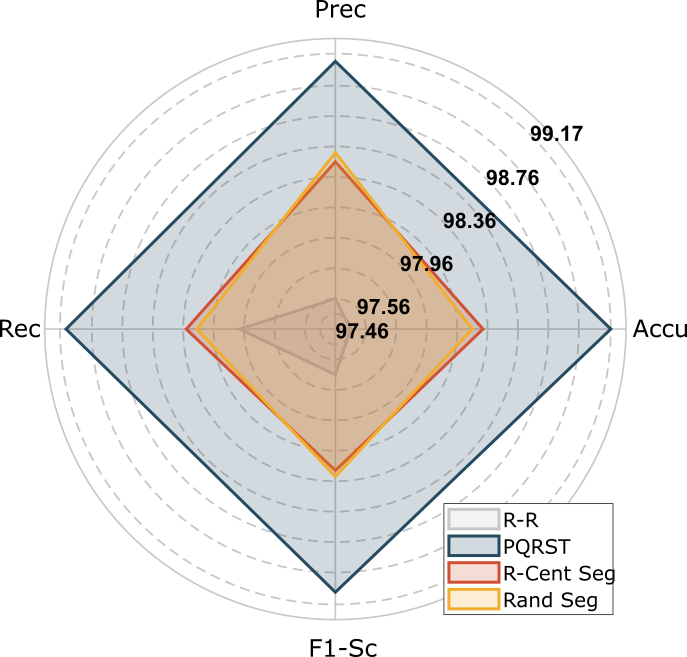}}\hfill
\subfloat[PTB Dataset]{\includegraphics[width=0.33\textwidth]{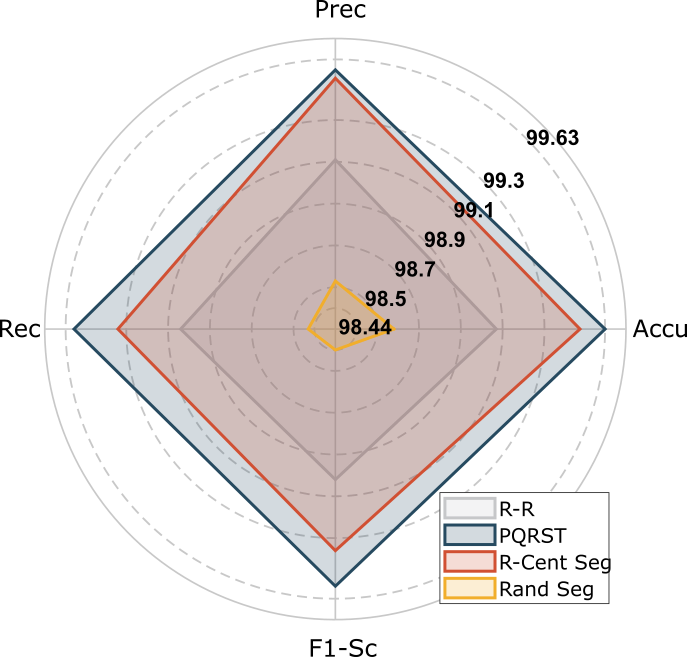}}\\
\caption*{(B) Inception-Time CNN Architecture}
\end{figure}

\begin{figure}[H]\ContinuedFloat
\subfloat[ECG-ID Dataset]{\includegraphics[width=0.31\textwidth]{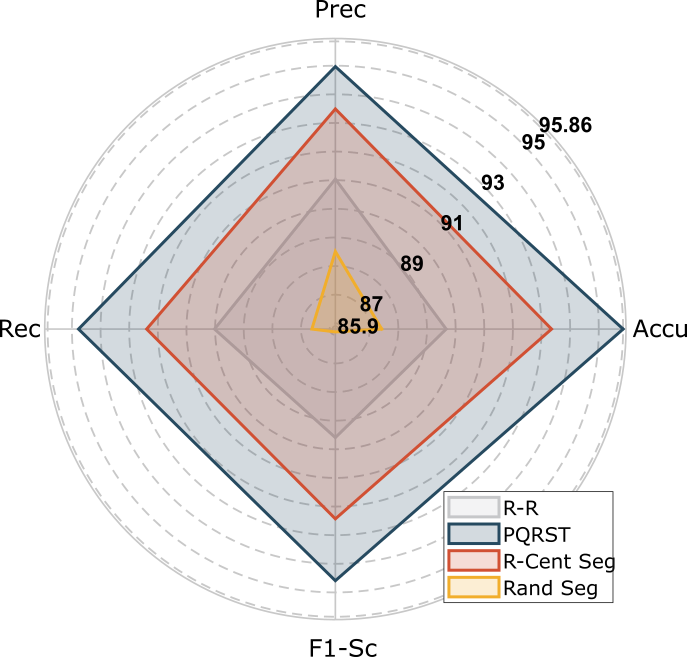}}\hfill
\subfloat[MIT-BIH Dataset]{\includegraphics[width=0.31\textwidth]{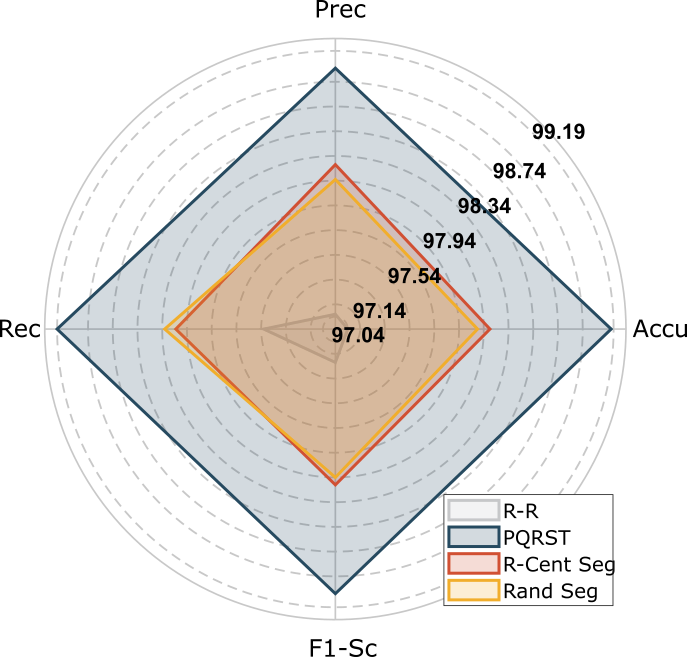}}\hfill
\subfloat[PTB Dataset]{\includegraphics[width=0.31\textwidth]{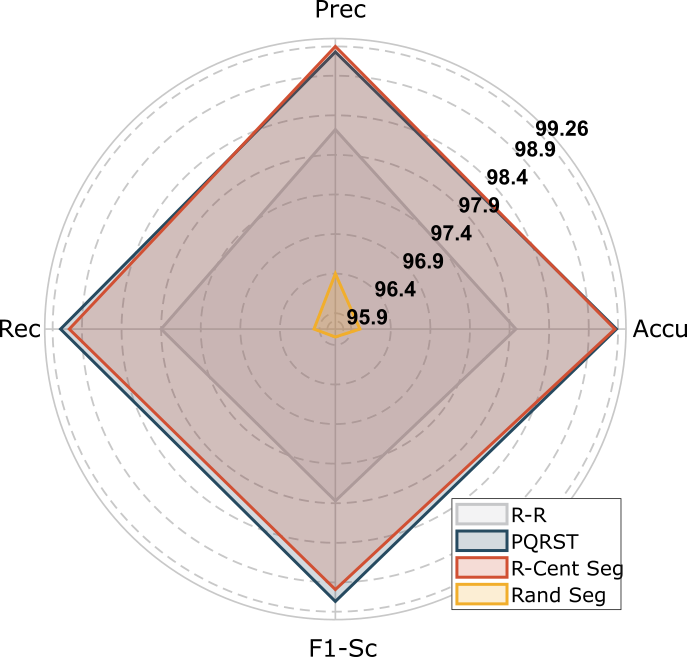}}\\
\caption*{(C) ECGNet CNN Architecture}
\end{figure}

\begin{figure}[H]\ContinuedFloat
\subfloat[ECG-ID Dataset]{\includegraphics[width=0.31\textwidth]{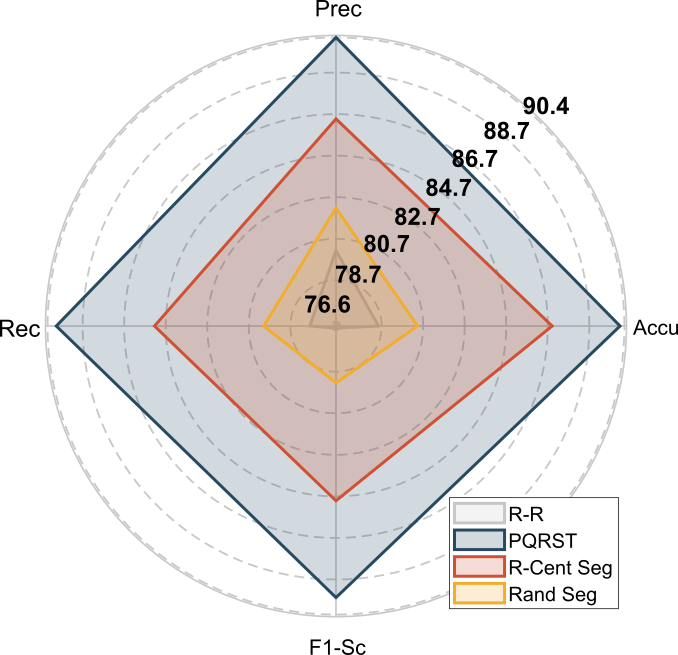}}\hfill
\subfloat[MIT-BIH Dataset]{\includegraphics[width=0.31\textwidth]{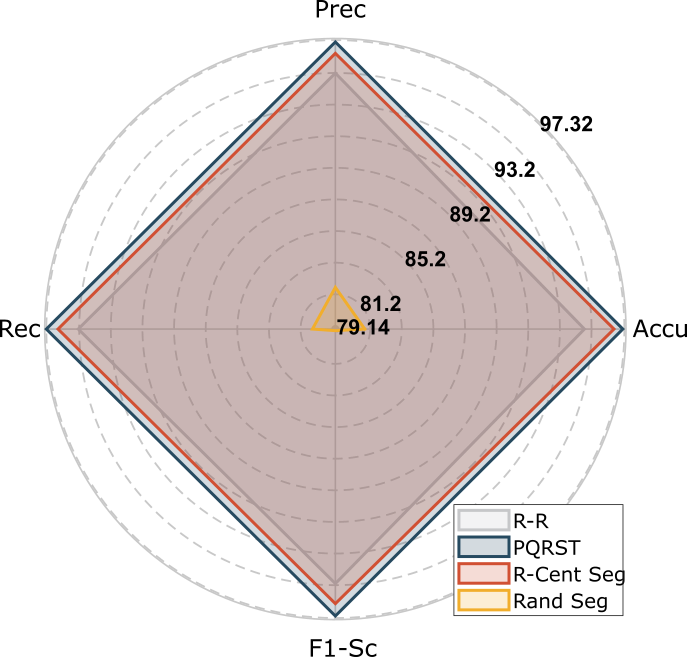}}\hfill
\subfloat[PTB Dataset]{\includegraphics[width=0.31\textwidth]{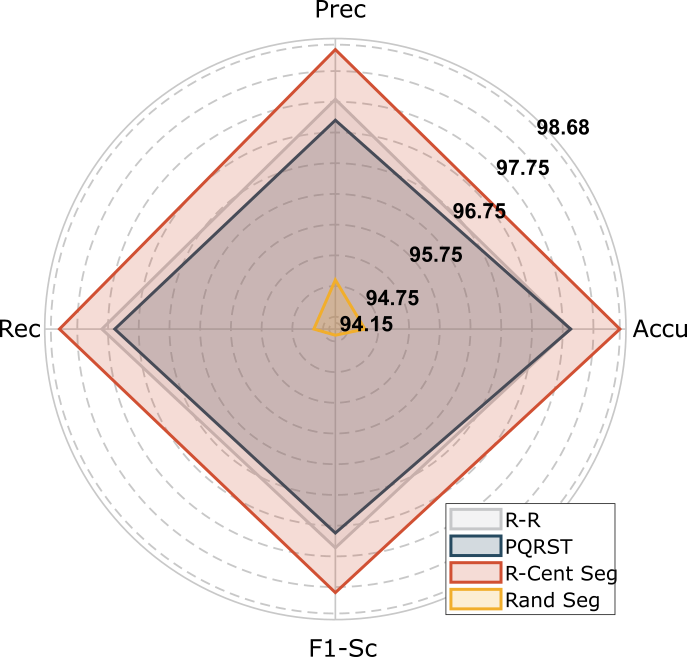}}\\
\caption*{(D) XCM CNN Architecture}

\caption{Radar charts for 1D CNN architectures across different segmentation strategies and evaluation metrics on the ECG-ID, MIT-BIH, and PTB datasets. Each radar chart uses adaptive scaling to the performance range of the specific dataset and model configuration.}
\label{1dcomp}
\end{figure}

The key observations from the depicted evaluation include:

\begin{itemize}
  \item The P–T segmentation strategy, which preserves the complete morphological structure of the cardiac cycle (P-QRS-T), consistently yielded superior performance across nearly all model–dataset combinations. This suggests that retaining the full electrophysiological context is critical for maximizing feature discriminability in 1D CNNs.
  \item Minor exceptions were noted for the XCM and ResNet-1D models on the PTB dataset, where R-centered segmentation marginally outperformed P–T segmentation. This may be attributable to PTB’s clinical origin, where certain pathologies affect T-wave morphology, making the QRS complex a more stable biometric feature.
  \item Both R–R interval and random segmentation strategies resulted in significantly degraded performance across all architectures, as indicated by their substantially reduced radar chart coverage. This confirms that rhythm-based or arbitrary segmentation fails to capture sufficient morphological detail for reliable identity recognition.
  \item Regarding architectural performance, InceptionTime combined with P–T segmentation achieved the highest overall biometric accuracy, attaining a mean weighted performance score of 98.63\% across all datasets. It was followed by ResNet-1D (98.05\%), ECGNet (97.76\%), and XCM (94.96\%) (Table \ref{tab:1dmean}). The superior performance of InceptionTime is likely due to its multi-branch design, which captures features at multiple temporal scales, a characteristic well-suited to the variable durations of P-QRS-T waves.
      
        \begin{table}[H]
        \centering
        \caption{Summary of Top-performing 1D CNN Architectures (Mean Score Across Metrics)}
        \label{tab:1dmean}
        \begin{tabular}{lllll}
        \hline
        Architecture  & ECG-ID & MIT-BIH & PTB    & Overall Mean \\ \hline

        InceptionTime & 97.2\% & 99.12\%  & 99.59\% & \textbf{98.63\%}      \\
        ResNet-1D     & 96.74\% & 98.24\%  & 99.16\% & 98.05\%      \\
        ECGNet        & 95.05\% & 99.09\%  & 99.14\% & 97.76\%      \\
        XCM           & 90.05\% & 97.23\%  & 97.59\% & 94.96\% \\   
        \hline 
        \end{tabular}
        \end{table}
      
\end{itemize}

In conclusion, the InceptionTime architecture with P–T wave segmentation was identified as the optimal 1D CNN configuration for ECG biometric recognition. This combination was therefore selected for subsequent feature fusion with 2D CNN models in the hybrid framework.

\subsection{2D CNNs results}

A subsequent comprehensive evaluation was conducted to assess the biometric recognition performance of four state-of-the-art 2D CNNs, including EfficientNetV2, ResNet-34, a Lightweight CNN, and a Vision Transformer (ViT), across the same set of ECG segmentation strategies, as detailed in Section \ref{sub3.4.2}. Each architecture was evaluated using various signal segmentation strategies derived from: (1) P–T segments; (2) R-peak-centered segments; (3) R–R intervals; and (4) random fixed-length segments.

\medskip

The models were evaluated on the same three datasets (ECG-ID, MIT-BIH, PTB) under an identical multi-metric protocol to ensure a direct and fair comparison with the 1D CNN results. The results are synthesized in Figure \ref{2dcomp} using multi-axis radar charts with adaptive scaling, providing a holistic visualization of each model's performance profile across the different segmentation types. Direct quantitative comparisons between charts should be made using the numerical results reported in Table \ref{tab:2dmean}.

\begin{figure}[H]
\centering
\subfloat[ECG-ID Dataset]{\includegraphics[width=0.31\textwidth]{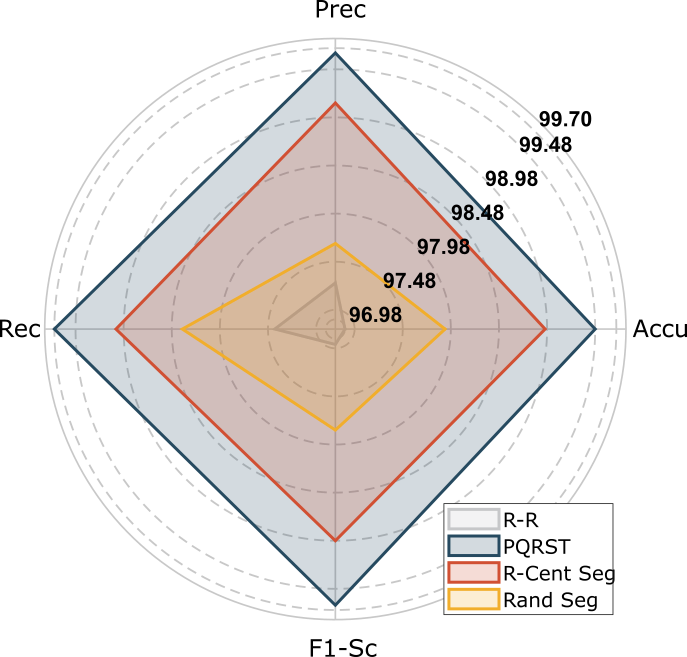}}\hfill
\subfloat[MIT-BIH Dataset]{\includegraphics[width=0.31\textwidth]{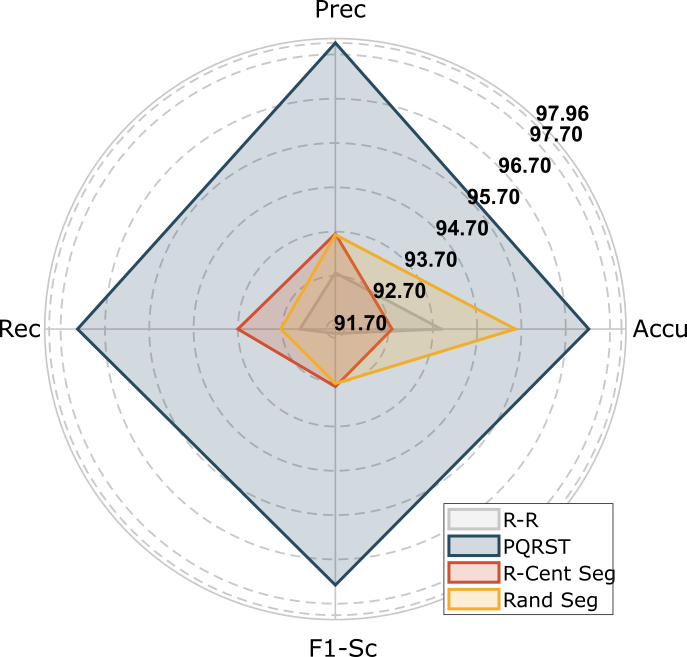}}\hfill
\subfloat[PTB Dataset]{\includegraphics[width=0.31\textwidth]{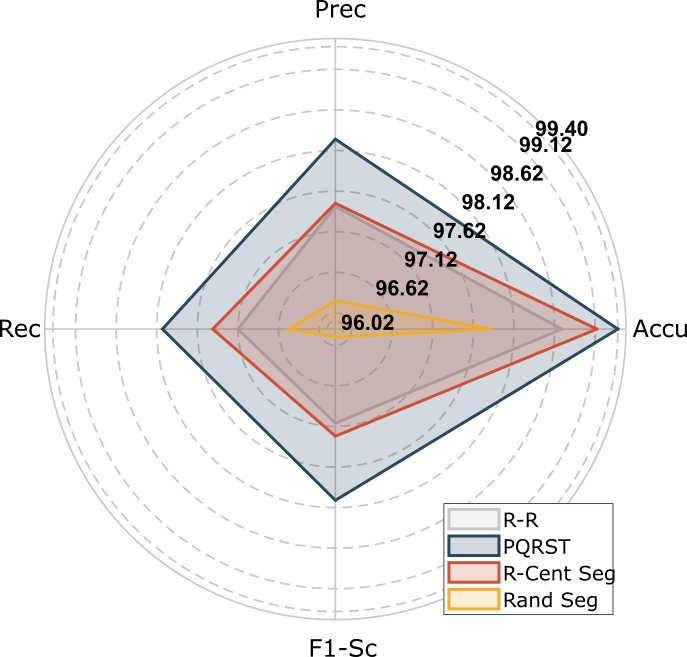}}\\
\caption*{(A) EfficientNetV2 CNN Architecture}
\end{figure}

\begin{figure}[H]\ContinuedFloat
\subfloat[ECG-ID Dataset]{\includegraphics[width=0.31\textwidth]{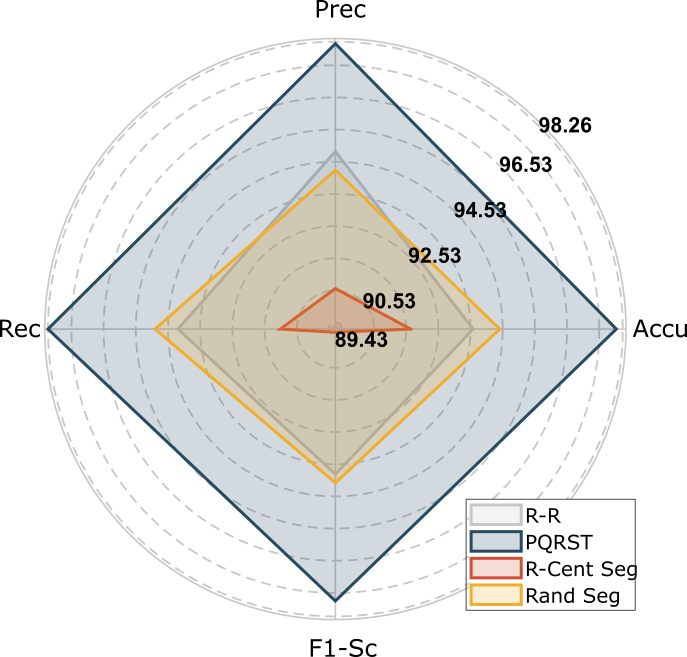}}\hfill
\subfloat[MIT-BIH Dataset]{\includegraphics[width=0.31\textwidth]{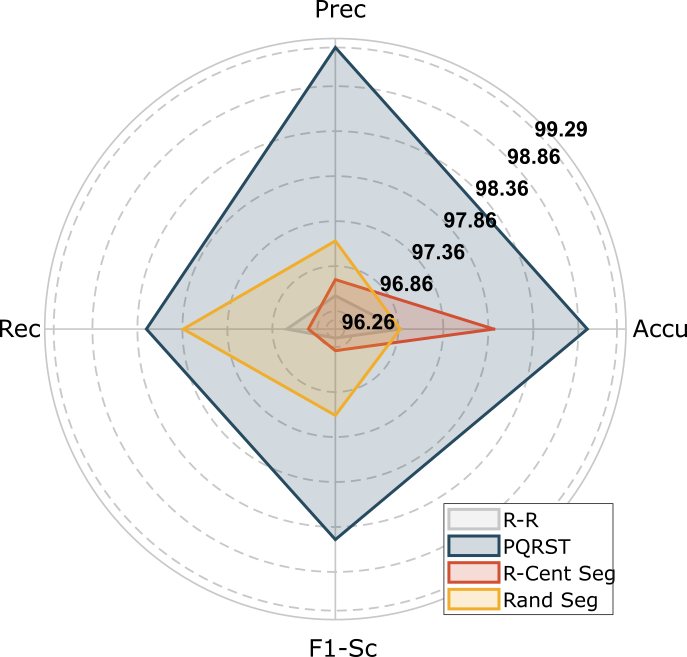}}\hfill
\subfloat[PTB Dataset]{\includegraphics[width=0.33\textwidth]{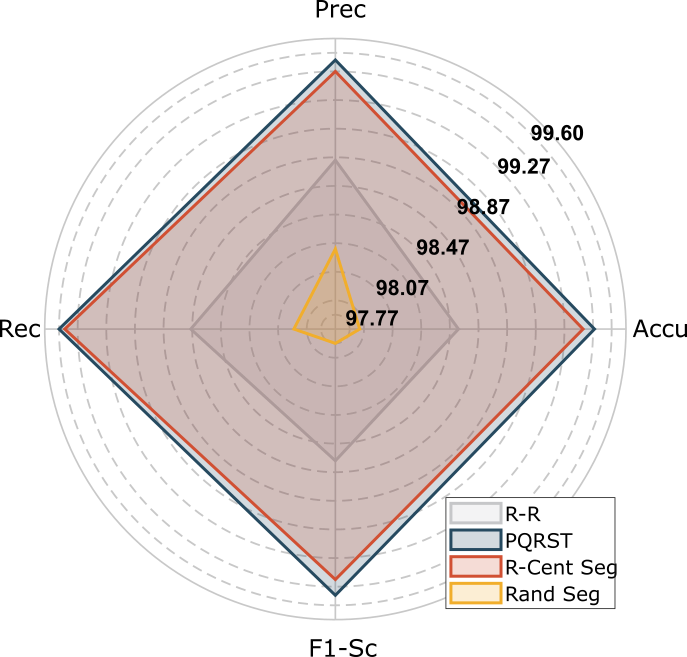}}\\
\caption*{(B) ResNet-34 CNN Architecture}
\end{figure}

\begin{figure}[H]\ContinuedFloat
\subfloat[ECG-ID Dataset]{\includegraphics[width=0.31\textwidth]{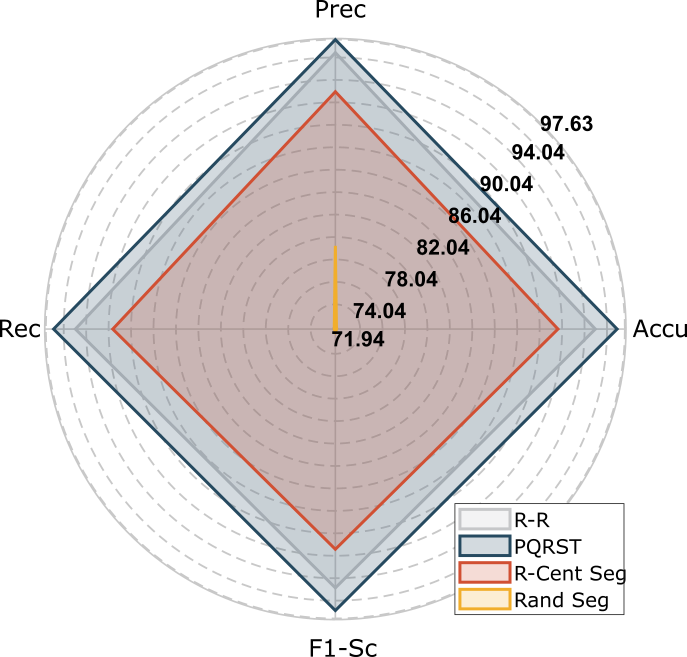}}\hfill
\subfloat[MIT-BIH Dataset]{\includegraphics[width=0.31\textwidth]{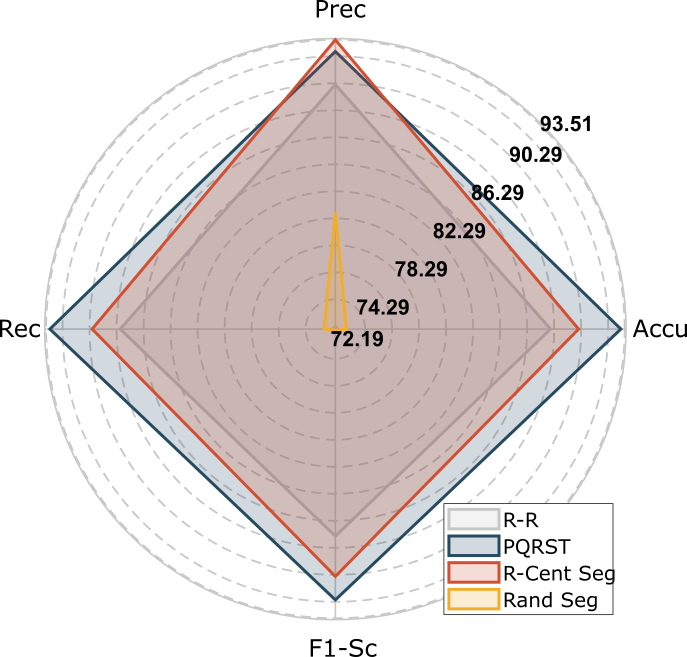}}\hfill
\subfloat[PTB Dataset]{\includegraphics[width=0.31\textwidth]{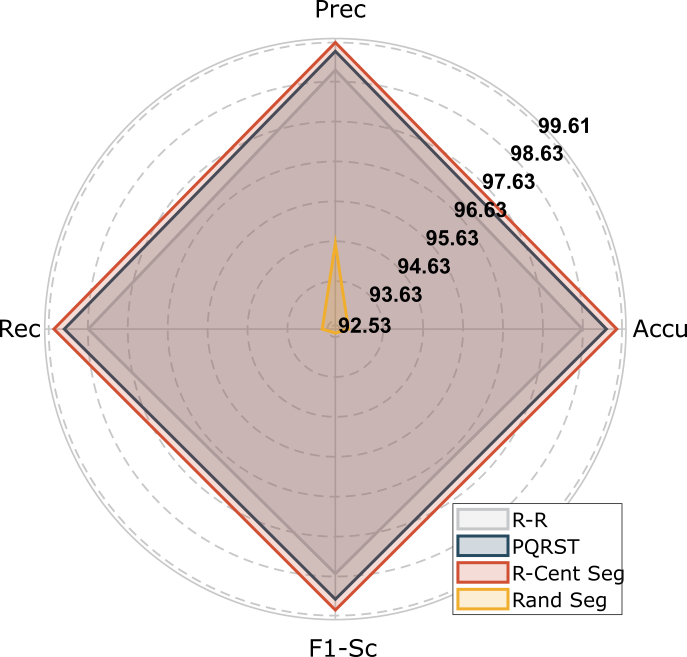}}\\
\caption*{(C) Lightweight CNN Architecture}
\end{figure}

\begin{figure}[H]\ContinuedFloat
\subfloat[ECG-ID Dataset]{\includegraphics[width=0.31\textwidth]{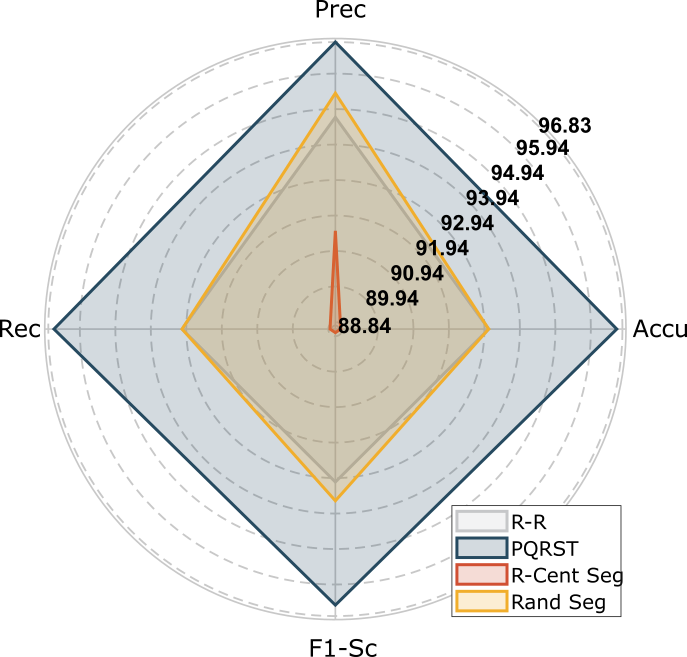}}\hfill
\subfloat[MIT-BIH Dataset]{\includegraphics[width=0.31\textwidth]{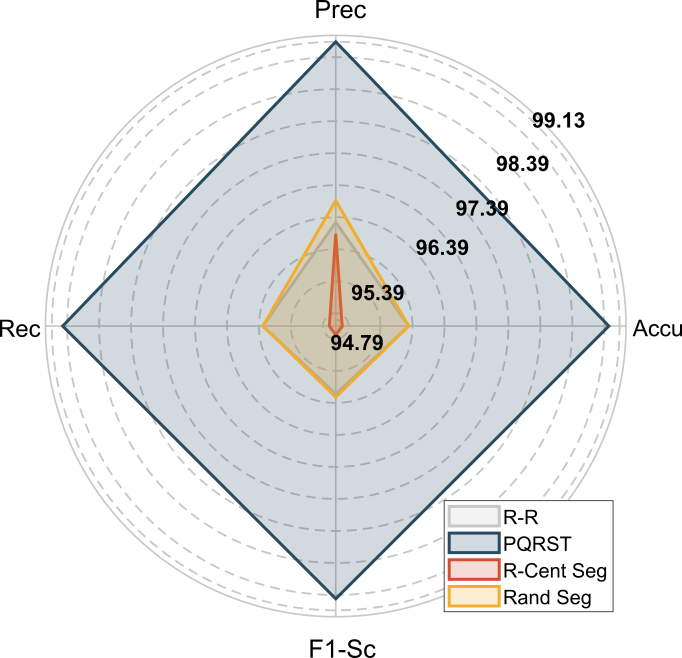}}\hfill
\subfloat[PTB Dataset]{\includegraphics[width=0.31\textwidth]{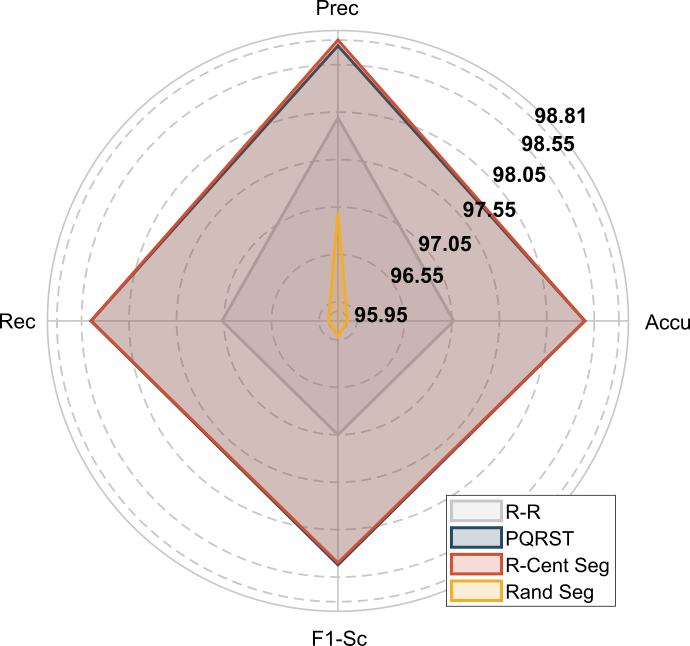}}\\
\caption*{(D) ViT CNN Architecture}
 
\caption{Radar charts for 2D CNN architectures (EfficientNetV2, ResNet-34, Lightweight CNN, ViT) across different segmentation strategies and evaluation metrics on the ECG-ID, MIT-BIH, and PTB datasets. Each radar chart uses adaptive scaling to the performance range of the specific dataset and model configuration.}
\label{2dcomp}
\end{figure}

Based on the results illustrated in the radar charts, several key observations can be deduced:

\begin{itemize}
  \item The P–T segmentation strategy again emerged as the most effective overall, consistently yielding the highest performance across nearly all model-dataset combinations. This underscores that providing the complete cardiac cycle as input, thereby retaining the full spectral signature of the P-QRS-T complex in the time-frequency domain, is paramount for optimal feature extraction by 2D CNNs.
      
  \item A singular exception was noted for the Lightweight CNN on the PTB dataset, where R-centered segmentation marginally surpassed the P–T segment. This anomaly may be attributed to the model's limited parameter capacity, which might struggle to process the more complex, longer P–T scalograms effectively, especially when confronted with the pathological morphologies present in PTB, making the simpler, focused QRS complex a more reliable feature.

  \item Mirroring the 1D CNN results, both R–R interval and random segmentation strategies resulted in the poorest performance across all architectures. This confirms that these strategies fail to provide a coherent or informative time-frequency representation for the network to learn discriminative features.
      
  \item Regarding architectural performance, ResNet-34 paired with P–T segmentation achieved the highest overall biometric accuracy, attaining a mean weighted performance score of 98.97\% across all datasets (Table \ref{tab:2dmean}). It was followed by EfficientNetV2 (98.50\%), ViT (98.06\%), and the Lightweight CNN (96.40\%). The superior performance of ResNet-34 can be attributed to its deep residual architecture, which is particularly effective at learning hierarchical features from image-like representations without submitting to degradation issues, making it exceptionally well-suited for interpreting complex scalograms.
      
        \begin{table}[H]
        \centering
        \caption{Summary of Top-performing 2D CNN Architectures (Mean Score Across Metrics)}
        \label{tab:2dmean}
        \begin{tabular}{lllll}
        \hline
        Architecture  & ECG-ID & MIT-BIH & PTB    & Overall Mean \\ \hline

        ResNet-34 & 98.08\% & 98.75\%  & 99.54\% & \textbf{98.97\%}      \\
        EfficientNetV2     & 99.62\% & 97.45\%  & 98.44\% & 98.50\%      \\
        ViT        & 96.67\% & 99\%  & 98.52\% & 98.06\%      \\
        Lightweight CNN           & 97.11\% & 92.82\%  & 99.26\% & 96.40\% \\   
        \hline 
        \end{tabular}
        \end{table}  

\end{itemize}

In conclusion, the ResNet-34 architecture with P–T wave segmentation was identified as the optimal 2D CNN configuration for ECG biometric recognition using time-frequency representations. This combination was therefore selected for the subsequent hybrid fusion with the leading 1D CNN model (InceptionTime).

\subsection{Ablation analysis of fusion strategies}

To rigorously evaluate the efficacy of the proposed multimodal integration framework, an ablation study was conducted comparing three distinct fusion methodologies: score-level fusion, feature-level fusion, and an attention-based fusion mechanism. This analysis was performed across the ECG-ID, MIT-BIH, and PTB datasets to determine the quantitative improvement gained by combining 1D (temporal) and 2D (time-frequency) ECG representations for biometric authentication, thereby isolating the contribution of the fusion mechanism itself. The fused models were benchmarked against the top-performing unimodal baselines: InceptionTime (for 1D raw signals) and ResNet-34 (for 2D scalogram representations). The implementation of each fusion strategy differed in its handling of integrative parameters:

\begin{itemize}
  \item Feature-level fusion operates by concatenating the high-dimensional feature vectors extracted from the penultimate layers of the 1D and 2D networks. This approach requires no tunable hyperparameters for the fusion itself, as it relies on the subsequent classification layers to learn the optimal combination of features during training.
  \item Score-level fusion combined the softmax outputs from the two unimodal models using a fixed weighting parameter, $\lambda$. A comprehensive sweep of $\lambda$ values (0.1, 0.3, 0.5, 0.7, 0.9) was performed to determine the optimal contribution balance (Figure \ref{aclam}).
      
      \begin{figure}[H]
        \centering
        \includegraphics[width=0.4\linewidth]{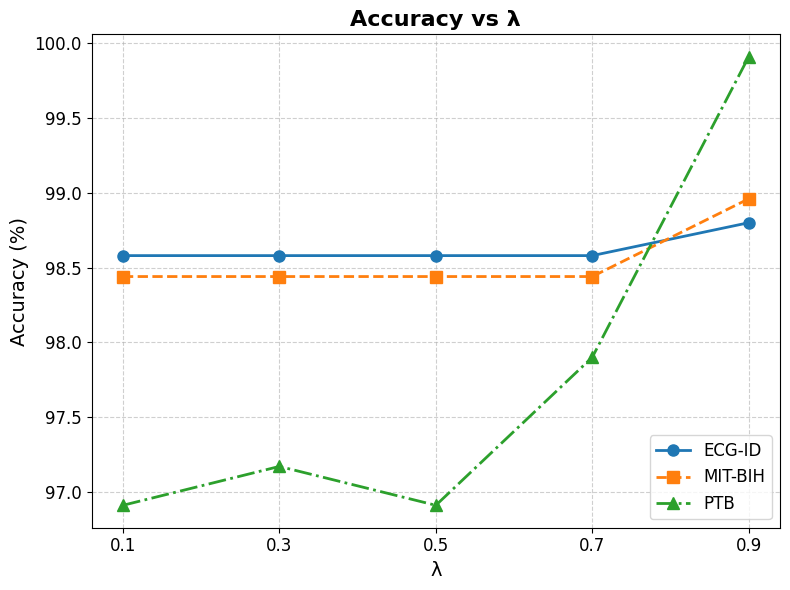}
        
        \caption{Accuracy of the score-level fusion model across different fusion weights \(\lambda\) for the ECG-ID, MIT-BIH, and PTB datasets. A parameter sweep over \(\lambda \in [0.1, 0.9]\) in increments of 0.1 identified \(\lambda = 0.9\) as optimal across all datasets, achieving accuracies of 98.80\% (ECG-ID), 98.96\% (MIT-BIH), and 99.91\% (PTB). The consistent optimal weight \(\lambda = 0.9\) indicates that the 2D spectral features contribute more significantly to the fused decision.}\label{aclam}
      \end{figure}
      
  \item Attention-based fusion introduces a learnable, adaptive weighting mechanism. Instead of a fixed value, the weighting fusion coefficient $\alpha$ is parameterized and optimized end-to-end via backpropagation. This allows the model to dynamically calibrate the contribution of each modality on a per-input basis, potentially improving adaptability and overall performance.
\end{itemize}

This structured ablation framework allows for a clear comparison between static, parameter-free, and intelligent, adaptive fusion schemes, directly measuring the value of learned integration for enhancing biometric authentication performance. Analysis of Figure \ref{aclam} indicates that a score-level fusion weight of $\lambda = 0.9$ yielded peak performance across all datasets, achieving accuracies of 98.80\% (ECG-ID), 98.96\% (MIT-BIH), and 99.91\% (PTB). This suggests that time-frequency features extracted by the 2D CNN provide a more robust foundation for classification, warranting a greater contribution in the fused decision.

\medskip

\begin{figure}[H]
\centering
\subfloat[ECG-ID Dataset]{\includegraphics[width=0.50\textwidth]{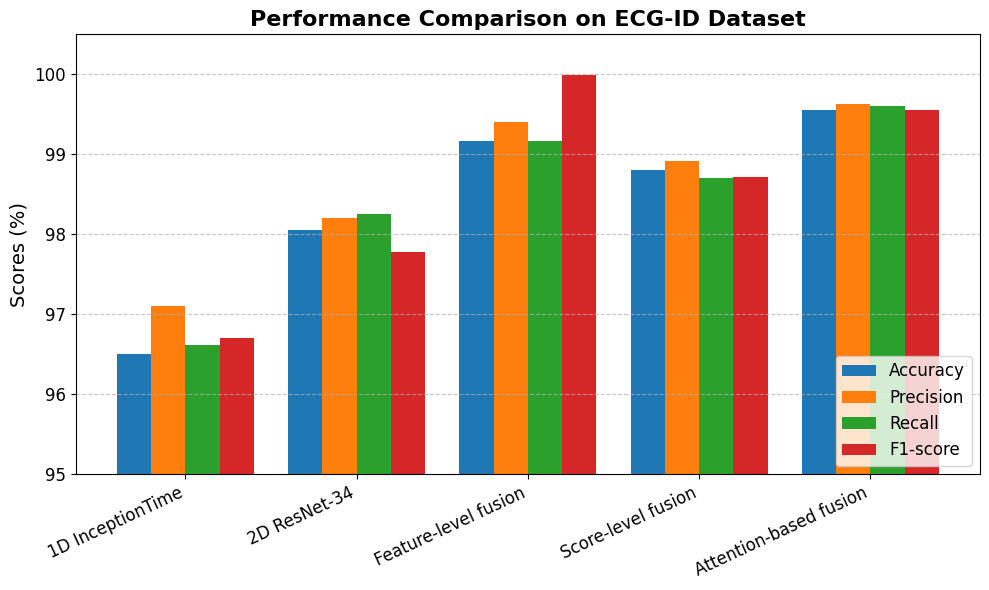}}\hfill
\subfloat[MIT-BIH Dataset]{\includegraphics[width=0.50\textwidth]{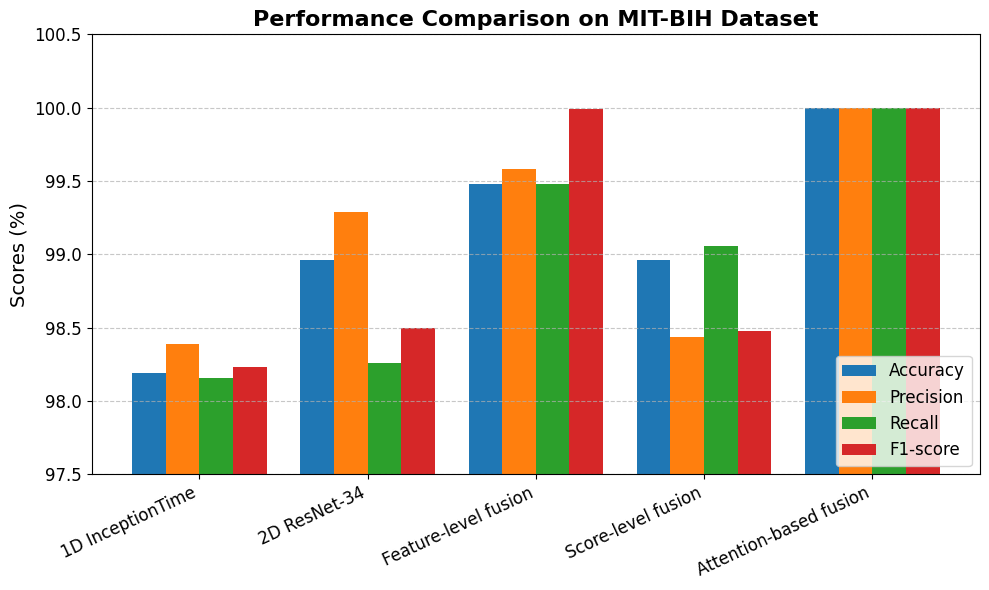}}\vfill
\subfloat[PTB Dataset]{\includegraphics[width=0.50\textwidth]{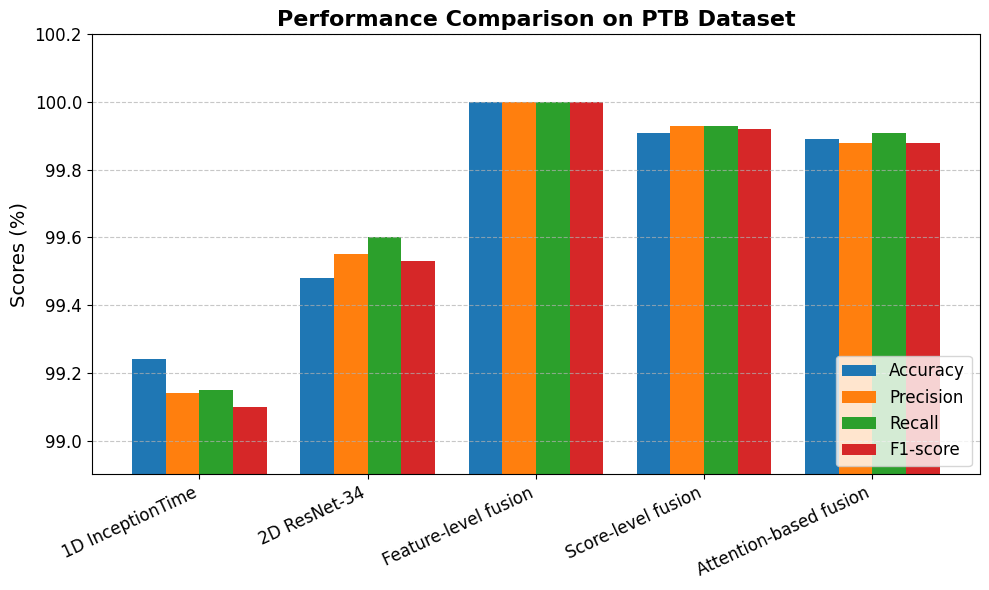}}

\caption{Comparative performance of unimodal baselines (InceptionTime and ResNet-34) and multimodal fusion strategies (feature-level, score-level, and attention-based) on the (a) ECG-ID, (b) MIT-BIH, and (c) PTB datasets. Performance is measured by accuracy (\%). The attention-based fusion mechanism achieves the highest accuracy across all datasets, demonstrating the effectiveness of adaptive modality weighting.}\label{ablfuz}
\end{figure}

The results of the ablation study are visually summarized in Figure \ref{ablfuz} using bar charts, which provide a clear comparison of the authentication performance across the different fusion strategies and the unimodal baselines. This graphical representation highlights the relative effectiveness of each approach and illustrates the performance gains achieved through multimodal integration compared to the standalone 1D and 2D models.

The bar charts in Figure \ref{ablfuz} reveal several key findings from the ablation study:

\begin{itemize}
  \item The 2D CNN (ResNet-34) baseline consistently surpassed the 1D CNN (InceptionTime) baseline across all datasets and metrics. This confirms the superior discriminative power of time-frequency features derived from scalograms over raw temporal signals for ECG biometrics.
  \item All fusion strategies demonstrated a significant performance gain over the best unimodal model. This validates the central hypothesis that integrating complementary temporal and spectral features (combining both 1D and 2D representations) enhances recognition robustness and accuracy.
  \item The attention-based fusion mechanism emerged as the most effective strategy overall, achieving the highest accuracy on the ECG-ID (99.56\%) and MIT-BIH (100.00\%) datasets and a near-optimal 99.89\% on the PTB dataset (Table \ref{tab:optm}). The top performance achieved on the MIT-BIH dataset can be attributed to its limited number of classes compared to the other datasets. Its adaptive nature, which learned optimal modality weights of $\alpha = 0.623, 0.565$, and $0.508$ for the three datasets respectively, allowed it to dynamically prioritize the most informative features for each input, maximizing the synergy between the two modalities.
      \begin{table}[H]
      \centering
      \caption{Optimal Modality Weight ($\alpha$) and Final Accuracy of Attention-Based Fusion per Dataset}
      \label{tab:optm}
      \begin{tabular}{lll}
      \hline
      Dataset  & Optimal $\alpha$ & Accuracy \\ \hline

      ECG-ID & 0.623 & 99.56\%      \\
      MIT-BIH & 0.565 & 100.00\%      \\
      PTB & 0.508 & 99.89\%      \\
      \hline 
      \end{tabular}
      \end{table}
\end{itemize}

The ablation study confirms that a multimodal approach is strictly superior to unimodal analysis. Furthermore, it demonstrates that an intelligent, learnable fusion mechanism (attention-based) outperforms static fusion rules (score-level and feature-level), providing a robust and generalizable framework for high-accuracy ECG biometric recognition.

\subsection{Overfitting Analysis and Model Generalization}

To evaluate the generalization capability of the proposed framework and verify the absence of overfitting we present the following analyses. We report results for the best-performing configuration, namely the combination of InceptionTime, ResNet-34, and P–T wave segmentation with attention-based fusion. The reported results correspond to the optimal attention weights 
$\alpha=0.623$, $0.565$, and $0.508$ for the ECG-ID, MIT-BIH, and PTB datasets, respectively.

\begin{figure}[H]
\centering
\subfloat[ECG-ID Dataset]{\includegraphics[width=0.7\textwidth]{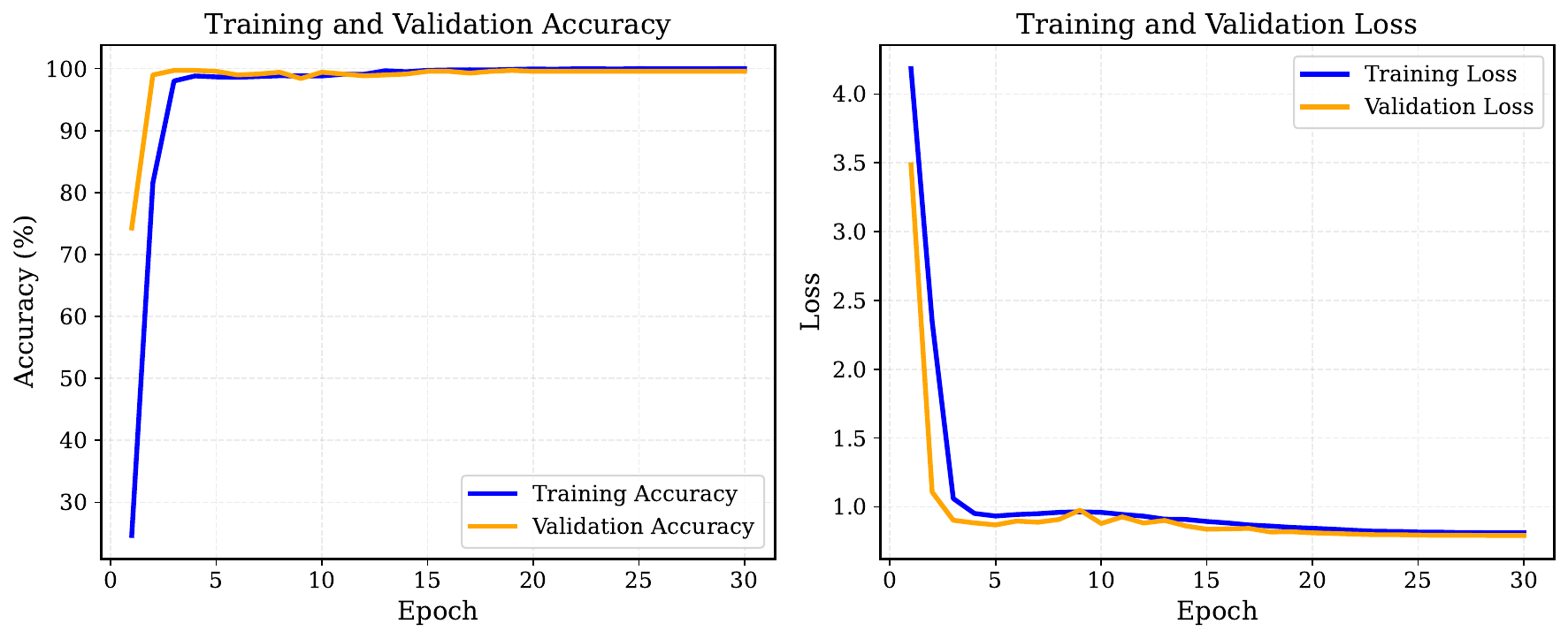}}\hfill
\end{figure}

\begin{figure}[H]\ContinuedFloat
\centering
\subfloat[MIT-BIH Dataset]{\includegraphics[width=0.7\textwidth]{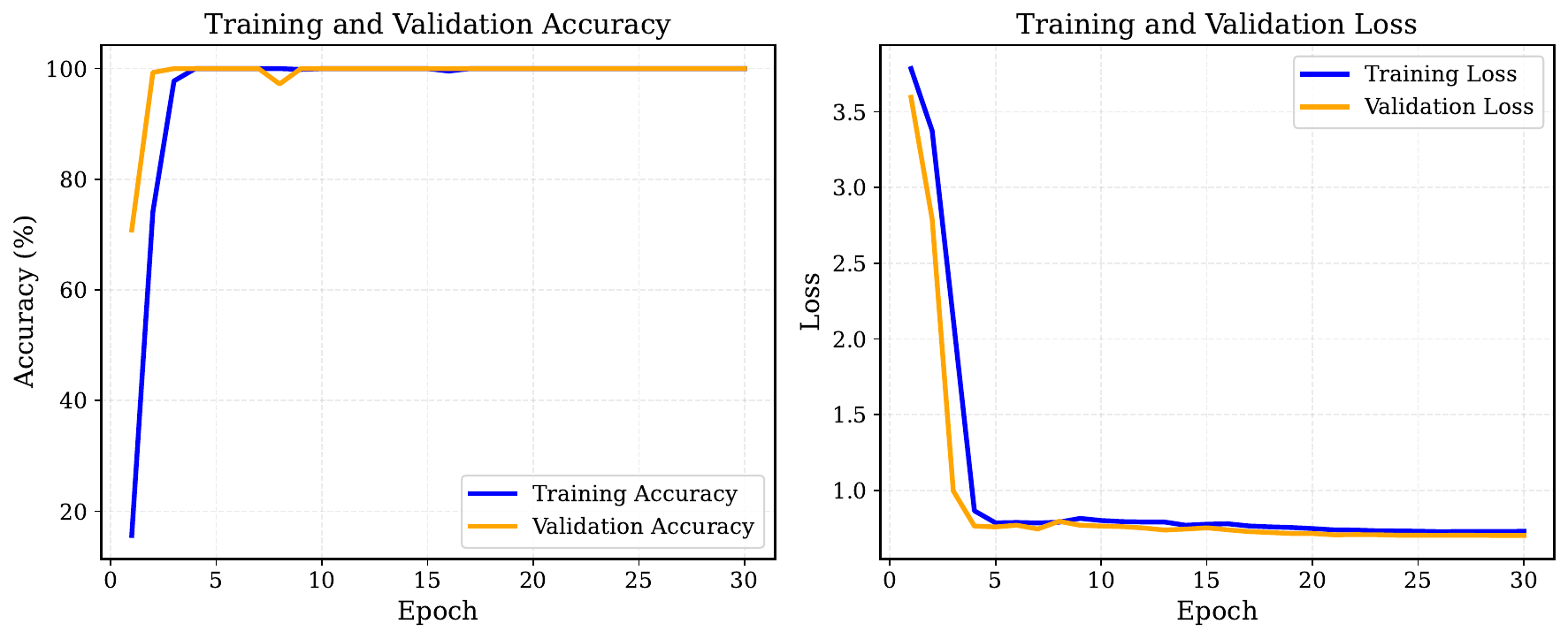}}\vfill
\end{figure}

\begin{figure}[H]\ContinuedFloat
\centering
\subfloat[PTB Dataset]{\includegraphics[width=0.7\textwidth]{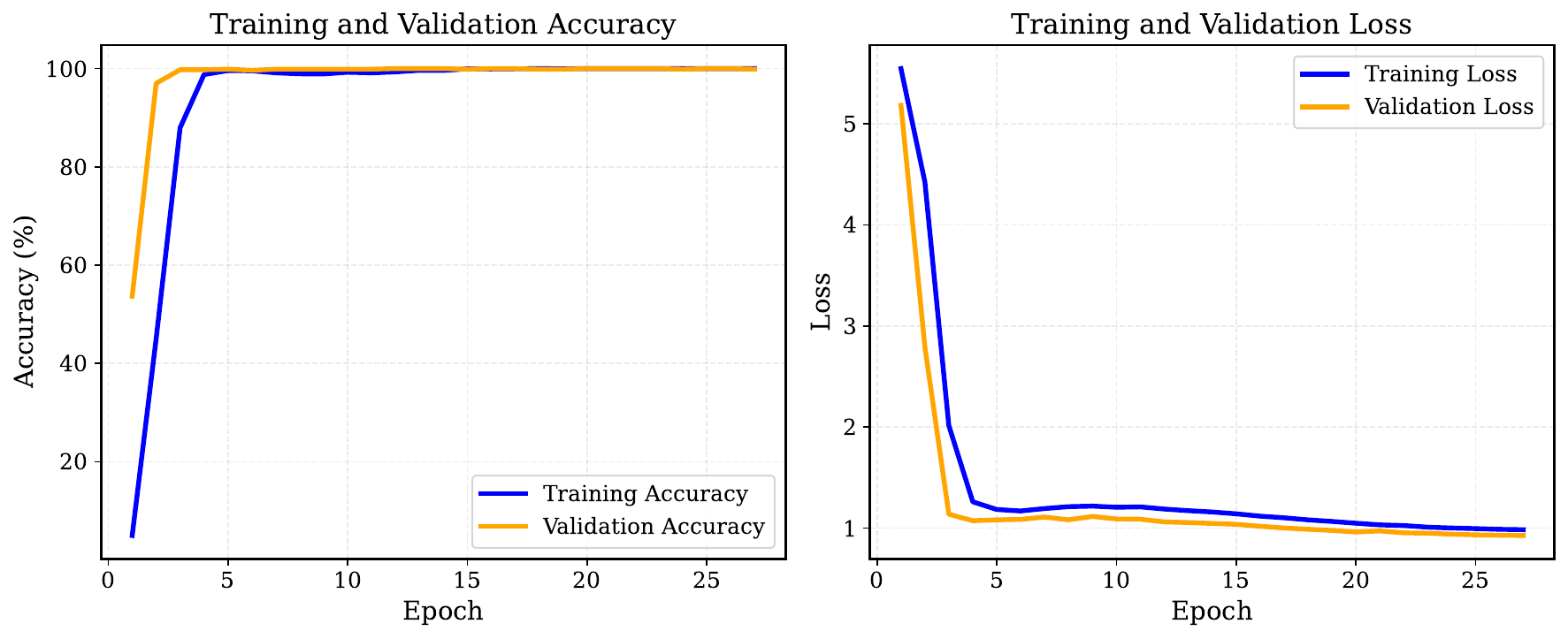}}
\caption{Training and validation learning curves (accuracy and loss) for the proposed attention-based fusion mechanism on (a) ECG-ID, (b) MIT-BIH, and (c) PTB datasets. The close alignment between training and validation curves, with a gap of less than 2\% at convergence, indicates stable learning and effective generalization without overfitting.}\label{fig:training_curves}
\end{figure}

Figure \ref{fig:training_curves} presents the training and validation learning curves (accuracy and loss) for the proposed attention-based fusion mechanism across all three datasets. The training and validation curves exhibit smooth and consistent convergence, with no significant divergence between them. Specifically, the validation accuracy closely tracks the training accuracy throughout training, with a gap of less than $2\%$ at convergence, while the validation loss decreases consistently without increasing in later epochs. This behavior indicates stable learning and good generalization, suggesting that the model does not suffer from overfitting and captures discriminative ECG patterns rather than dataset-specific noise.

\begin{figure}[H]
\centering
\subfloat[ECG-ID Dataset]{\includegraphics[width=0.33\textwidth]{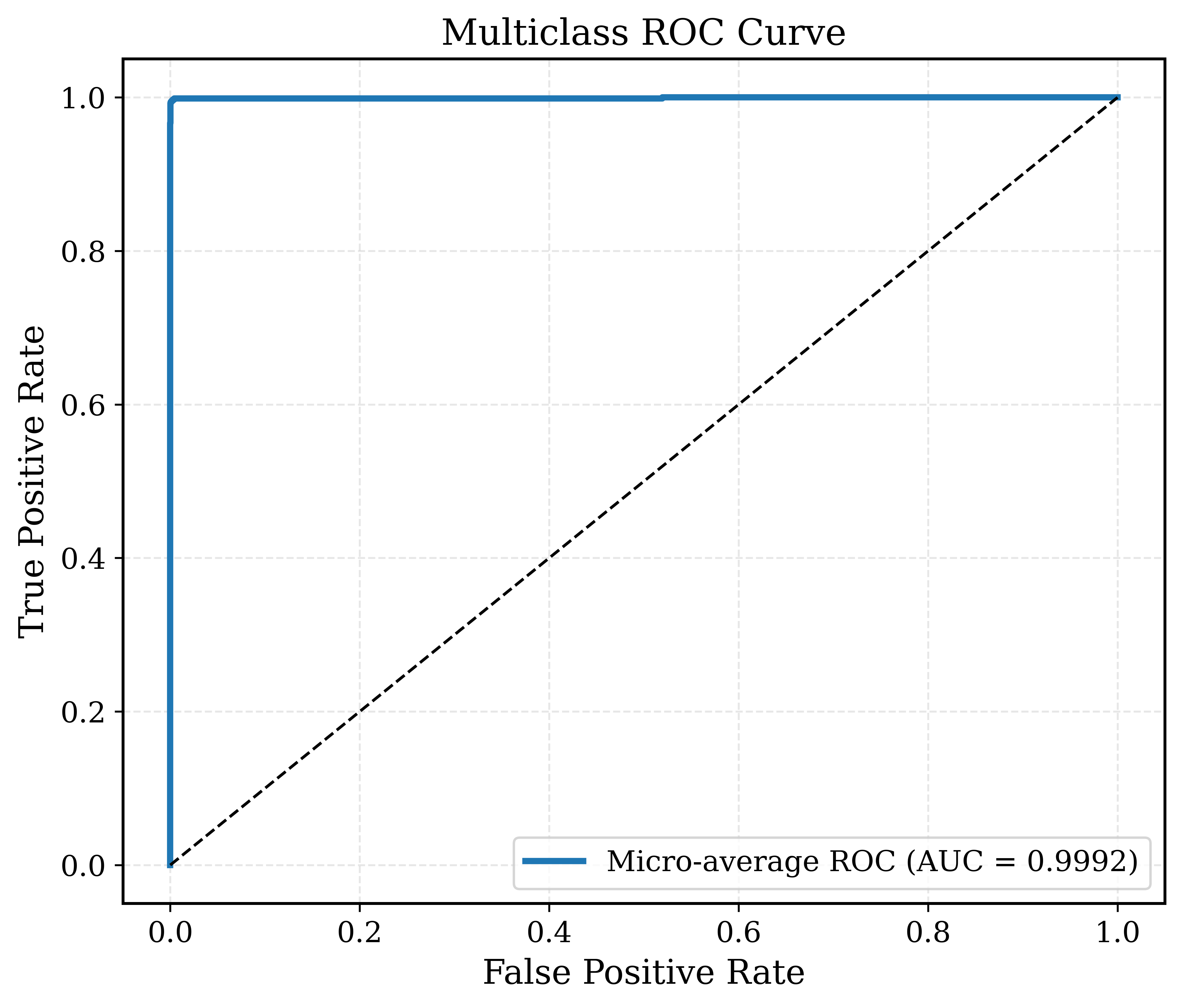}}\hfill
\subfloat[MIT-BIH Dataset]{\includegraphics[width=0.33\textwidth]{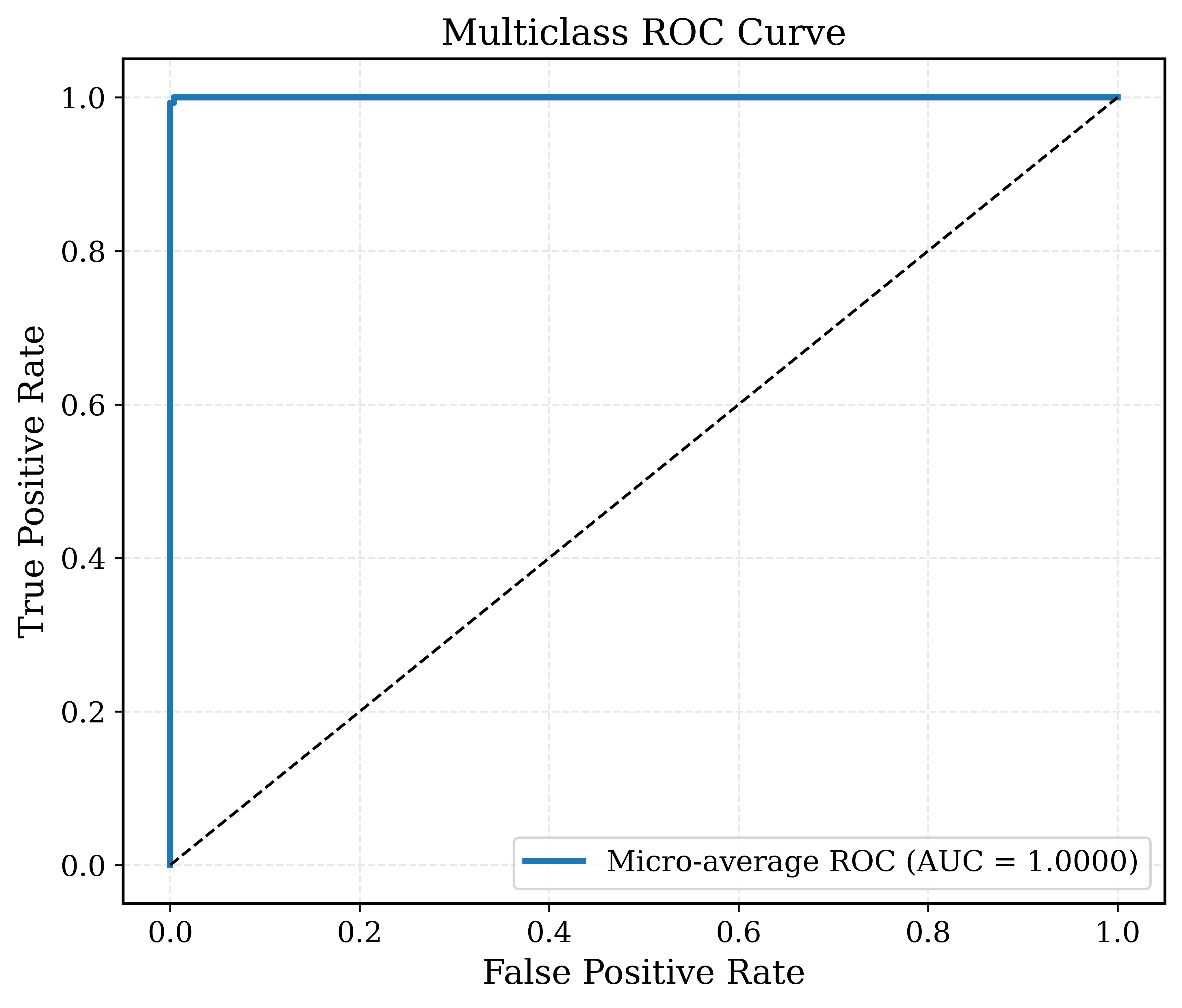}}\hfill
\subfloat[PTB Dataset]{\includegraphics[width=0.33\textwidth]{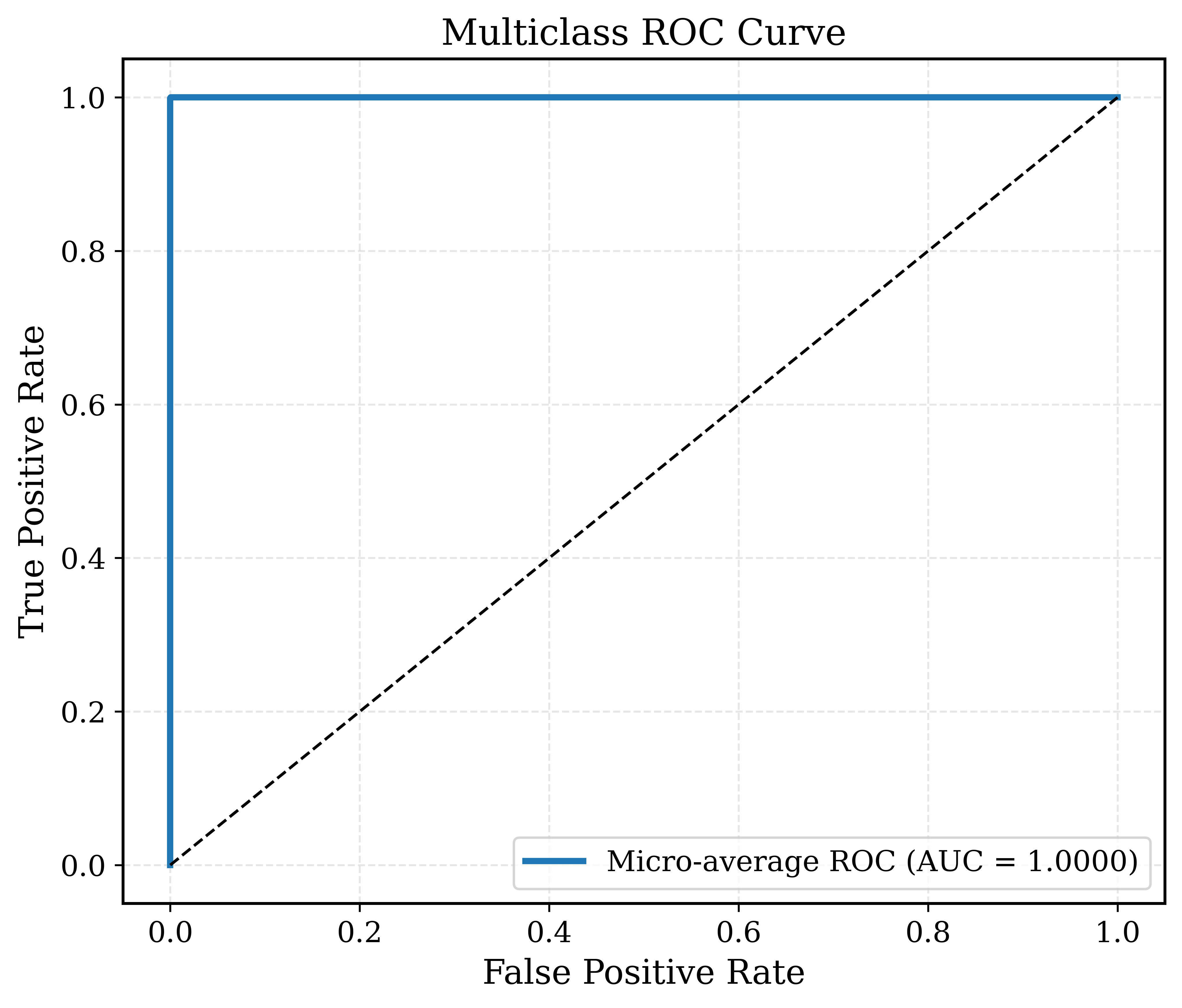}}
\caption{ROC curves with corresponding AUC values for the proposed attention-based fusion mechanism on (a) ECG-ID, (b) MIT-BIH, and (c) PTB datasets. The near-ideal AUC values (>0.997) demonstrate excellent discriminative capability.}\label{fig:roc_curves}
\end{figure}

The discriminative capability of the model is further confirmed by the Receiver Operating Characteristic (ROC) curves and corresponding Area Under the Curve (AUC) values presented in Figure \ref{fig:roc_curves}. The near-ideal AUC values, 0.9992 for ECG-ID, 1.0000 for MIT-BIH, and 1.0000 for PTB, demonstrate excellent separation between genuine and impostor matches, reflecting the model's strong discriminative power.

\begin{figure}[H]
\centering
\subfloat[ECG-ID Dataset]{\includegraphics[width=0.33\textwidth]{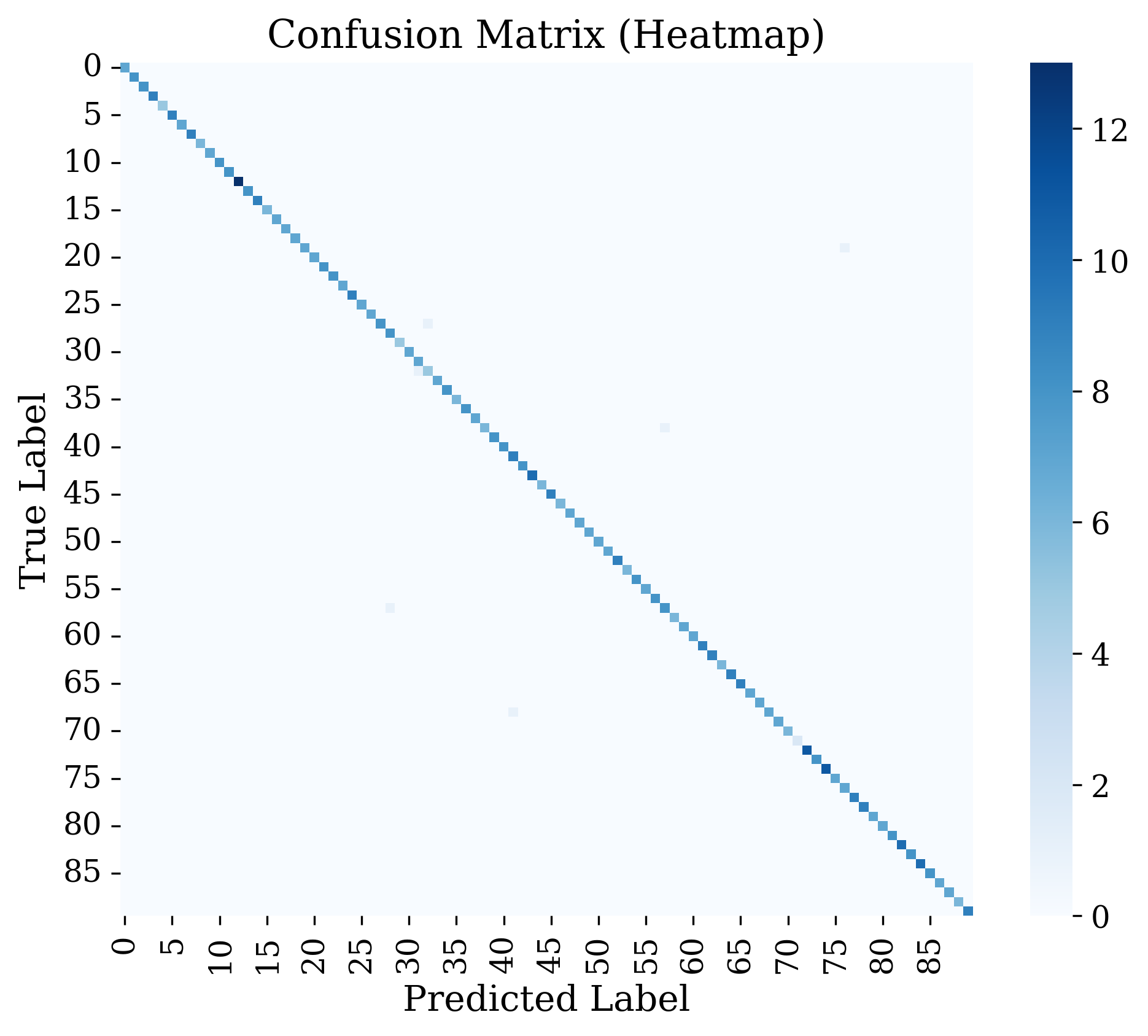}}\hfill
\subfloat[MIT-BIH Dataset]{\includegraphics[width=0.33\textwidth]{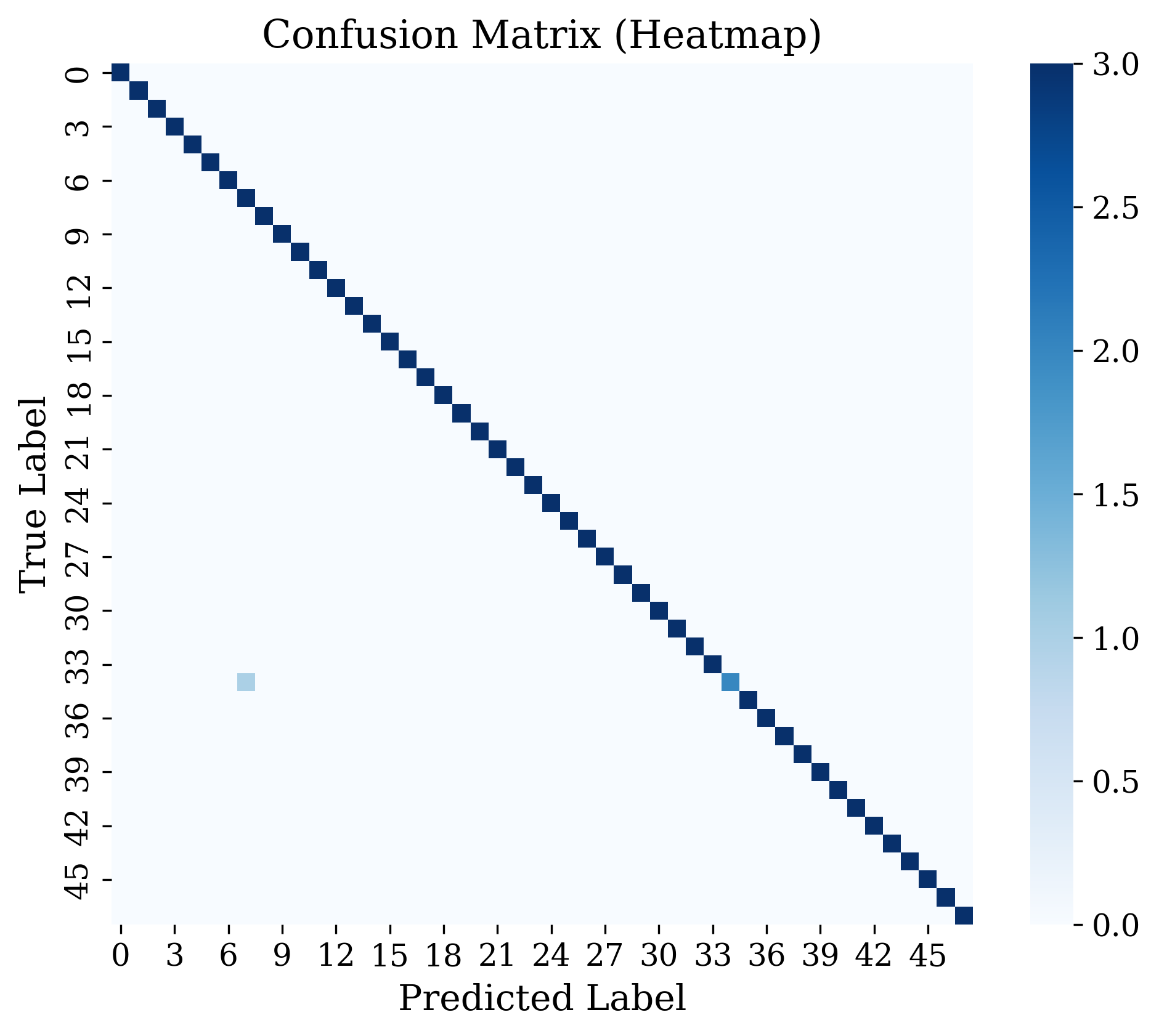}}\hfill
\subfloat[PTB Dataset]{\includegraphics[width=0.33\textwidth]{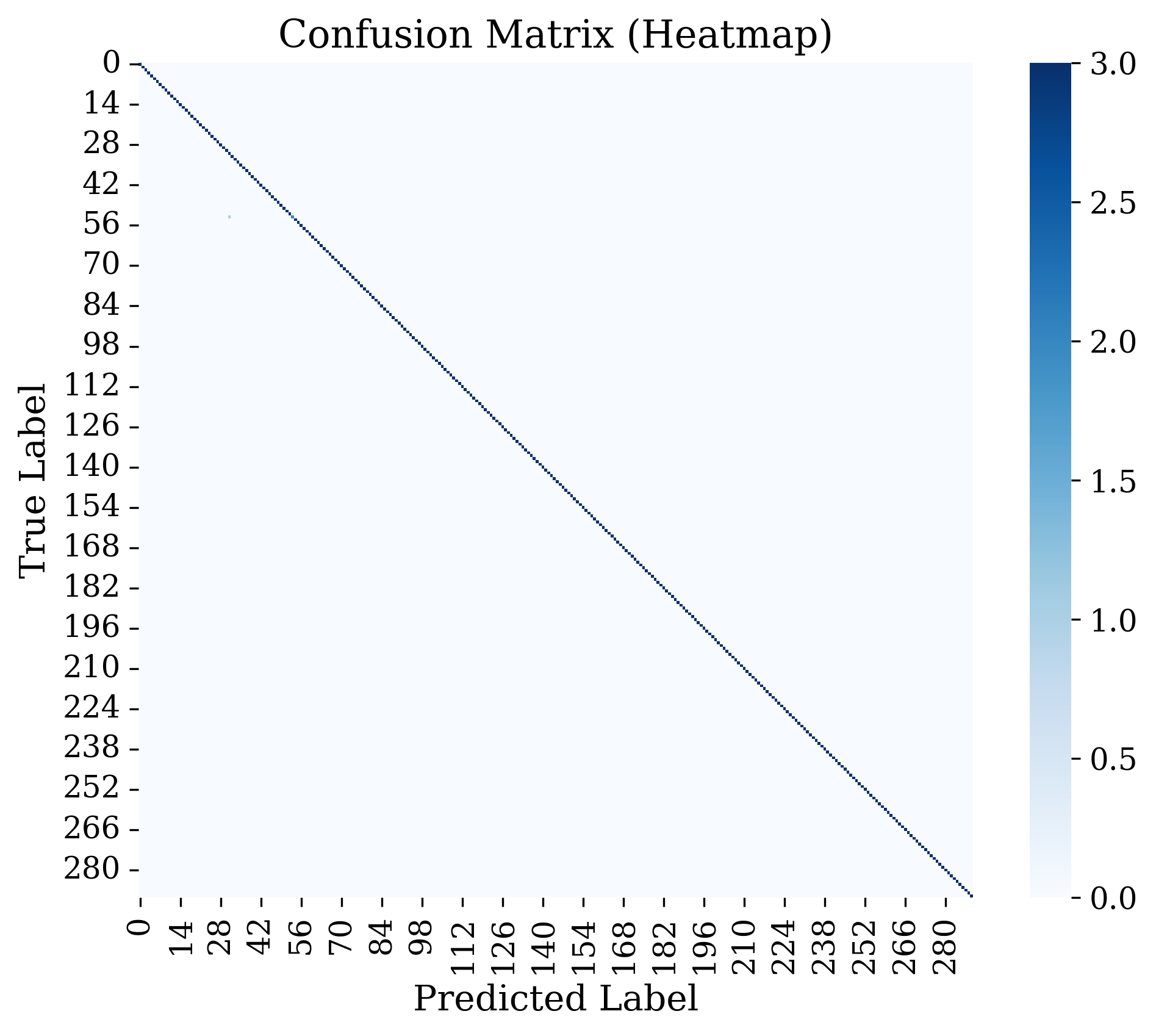}}
\caption{Confusion matrices for the proposed attention-based fusion mechanism on (a) ECG-ID, (b) MIT-BIH, and (c) PTB datasets. The strong concentration of correct predictions along the main diagonal confirms consistent classification performance across subjects.} \label{fig:confusion_matrices}
\end{figure}

Figure \ref{fig:confusion_matrices} presents the confusion matrices for all three datasets. The strong concentration of correct predictions along the main diagonal, with minimal misclassification between subjects, confirms consistent classification performance across subjects. The few off-diagonal errors are typically between subjects with similar morphological characteristics, which is expected in challenging biometric tasks.

\medskip

Overall, the combined analysis of learning curves, ROC-AUC, and confusion matrices confirms that the proposed framework achieves high performance while maintaining strong generalization across different datasets and experimental conditions. The close alignment of training and validation metrics, near-ideal ROC-AUC values, and consistently high diagonal concentration in confusion matrices provide compelling evidence that the reported results reflect genuine feature learning rather than overfitting to dataset-specific noise.

\subsection{Comparative analysis with state-of-the-art methods}

A comprehensive comparative analysis was conducted to benchmark the performance of our proposed 1D–2D CNN fusion model with an attention mechanism against recent state-of-the-art methods in ECG-based biometric recognition. The evaluation was performed under consistent experimental conditions using the same datasets (ECG-ID, MIT-BIH, and PTB) and evaluation protocols to ensure a fair comparison. The results, summarized in Table \ref{tab:comparison11}, demonstrate the superior performance of our multimodal fusion approach. Our model significantly outperformed existing 2D approaches, which primarily rely on time-frequency representations. For instance, on the MIT-BIH dataset, the previous best performance of 98.84\% achieved by Fuster-Barceló et al. \cite{fuster2022elektra} using EKM + CNN was surpassed by our method, which achieved 100.00\% accuracy. Similarly, on the PTB dataset, our model achieved 99.89\% accuracy, exceeding the 98.80\% accuracy reported by El Boujnouni et al. \cite{el2022wavelet}. Our approach also exceeded the performance of leading 1D methods, which process raw temporal signals. For example, Wang et al. \cite{wang2024ecg} achieved 99.15\% accuracy on PTB using a CNN with self-supervised contrastive learning, while our method attained 99.89\%. On MIT-BIH, Zehir et al. \cite{zehir2024empirical} reported 98.57\% accuracy with a GRU-LSTM architecture, whereas our model achieved perfect classification (100.00\%).

{\small
\begin{longtable}{
    >{\raggedright\arraybackslash}p{1.25cm} 
    >{\raggedright\arraybackslash}p{3cm} 
    c 
    >{\raggedright\arraybackslash}p{4.5cm} 
    c 
    c
    c 
}
\caption{Performance comparison with state-of-the-art ECG biometric methods. \\
\textbf{Note}: Dashes (–) indicate that the corresponding result was not reported in the original publication.}
\label{tab:comparison11}\\
\hline
\textbf{Class} & \textbf{Reference} & \textbf{Year} & \textbf{Method} & \multicolumn{3}{c}{\textbf{Accuracy (\%)}} \\
\cline{5-7}
 & & & & \textbf{ECG-ID} & \textbf{MIT-BIH} & \textbf{PTB} \\
\hline
\multirow{4}{*}{2D} & Ivanciu et al.~\cite{ivanciu2021ecg} & 2021 & Siamese Neural Networks & 86.47 & -- & -- \\
 & El Boujnouni et al.~\cite{el2022wavelet} & 2022 & CWT + DWT + Capsule Network & -- & 98.20 & 98.80 \\
 & Fuster-Barceló et al.~\cite{fuster2022elektra} & 2022 & EKM + CNN & -- & 98.84 & 89.80 \\
 & Al-Jibreen et al.~\cite{al2024person} & 2024 & Heart Beat Segmentation, CWT \& CNN & -- & 93.81 & -- \\
\hline
\multirow{6}{*}{1D} & Jyotishi and Dandapat~\cite{jyotishi2021ecg} & 2021 & LSTM & 93.11 & 96.81 & -- \\
 & Kim et al.~\cite{kim2023one} & 2023 & One-Dimensional Shallow Neural Network & 94.39 & 95.29 & 94.97 \\
 & Melzi et al.~\cite{melzi2023ecg} & 2023 & ECGXtractor & 97.75 & -- & -- \\
 & Wang et al.~\cite{wang2024ecg} & 2024 & CNN + Self-Supervised Contrastive Learning & 98.77 & 98.67 & 99.15 \\%98.25 & 97.96 & 98.67 \\
 & Zehir et al.~\cite{zehir2024empirical} & 2024 & GRU + LSTM & -- & 98.57 & 98.26 \\
 & Ma et al.~\cite{ma2024out} & 2024 & ORGNNFL & 92.57 & 98.88 & 97.80 \\
\hline
Fused 1D \& 2D & \textbf{Our approach} & \textbf{2025} & \textbf{1D--2D CNN Fusion with Attention Mechanisms} & \textbf{99.56} & \textbf{100.00} & \textbf{99.89} \\
\hline
\end{longtable}}

Overall performance, our proposed method achieved accuracies of 99.56\% on ECG-ID, 100.00\% on MIT-BIH, and 99.89\% on PTB, establishing a new state-of-the-art across all three datasets. These results underscore the effectiveness of integrating temporal (1D) and time-frequency (2D) features, with the attention mechanism further enhancing performance by dynamically optimizing the contribution of each modality based on input-specific characteristics.

\subsection{Multi-Session Analysis}

The datasets used in the previous experiments (i.e., ECG-ID, MIT-BIH, and PTB) involve recordings acquired within relatively limited time intervals for the same subjects. To further evaluate the proposed attention-guided fusion framework combining 1D and 2D CNNs, we additionally conducted experiments using the Heartprint dataset \cite{islam2022heartprint}. This dataset is particularly suitable for assessing long-term biometric stability, as it contains ECG recordings collected over extended periods, with the time interval between some sessions reaching up to 10 years. The objective of this analysis is to evaluate the uniqueness and temporal permanence of the proposed method, which are essential properties for reliable ECG-based biometric recognition.

\subsubsection{Description of the Heartprint Dataset}

The Heartprint dataset \cite{islam2022heartprint} is a large multi-session ECG biometric database comprising 1539 recordings collected from 199 healthy subjects. Each recording has a duration of 15 seconds and was acquired using finger-based ECG sensors under both resting and reading conditions. A key characteristic of this dataset is the long-term acquisition protocol, where signals were collected across multiple sessions spanning up to 10 years, with an average interval of 1572.2 days between the first session and the third session. In addition to the raw ECG signals, the dataset includes demographic information such as gender, ethnicity, and age groups. These characteristics make the Heartprint dataset a valuable benchmark for evaluating the uniqueness, robustness, and long-term permanence of ECG-based biometric recognition systems. Table \ref{tab:heartprint_summary} summarizes the characteristics of the multi-session Heartprint dataset.

\begin{table}[H]
\centering
\small

\caption{Summary of the characteristics of the multi-session Heartprint dataset}
\label{tab:heartprint_summary}
\begin{tabular}{|l|c|c|c|c|c|}
\hline
\textbf{Session} & \textbf{\#Subjects} & \textbf{\#Records} & \textbf{\#Records/Subject} & \textbf{Acquisition Date} & \textbf{Days from S1} \\
\hline
Session \#1 & 199 & 476 & 2--6 & January 2012 & -- \\
Session \#2 & 199 & 464 & 2--5 & June 2012 & 5--241 \\
Session \#3R & 109 & 365 & 3--6 & March 2022 & 36--3432 \\
Session \#3L & 78 & 234 & 3--3 & March 2022 & 71--3432 \\
\hline
\textbf{Total} & \textbf{199} & \textbf{1539} & \textbf{4--11} & -- & -- \\
\hline
\end{tabular}
\end{table}

\subsubsection{Evaluation Protocols and Results}

To assess the effectiveness of our proposed method, we followed the same three evaluation protocols conducted by the dataset's developers, namely: same-session, mixed-session, and cross-session protocols.

\begin{itemize}
  \item Same-Session Protocol: Training and testing are performed using data from the same acquisition session to evaluate baseline recognition capability. Experiments are conducted independently on each session (S1, S2, S3R, and S3L), where the training (80\%) and testing (20\%) sets are drawn from the same session.
  \item Mixed-Session Protocol: The training set consists of 80\% of the recordings from sessions S1 and S2, while the remaining 20\% of these recordings are used for testing. This protocol evaluates the model's ability to generalize when training and testing data originate from multiple sessions.
  \item Cross-Session Protocol: This protocol evaluates robustness across different acquisition sessions. The model is trained on data from one session (e.g., S1 or S2) and evaluated on data from a different session (e.g., S2, S3R, or S3L), assessing the impact of temporal variability and inter-session differences on recognition performance.
\end{itemize}

Table \ref{tab:heartprint_results} summarizes the evaluation protocols and reports the corresponding results obtained using the proposed attention-guided fusion framework. We consistently adopted the same experimental configuration as in previous sections, including Savitzky–Golay preprocessing, P–T wave segmentation, CWT-based transformation using the Morlet wavelet, InceptionTime for 1D feature extraction, ResNet-34 for 2D feature learning, and the attention-based fusion mechanism (Section 4.5).

\begin{table}[H]
\centering

\caption{Recognition performance of the proposed attention-guided 1D--2D CNN fusion framework on the Heartprint dataset under different evaluation protocols, compared with recently published methods}
\label{tab:heartprint_results}
\resizebox{\textwidth}{!}{%
\begin{tabular}{|l|l|l|c|c|c|c|c|}
\hline
\textbf{Protocol} & \textbf{Training Set} & \textbf{Testing Set} & \textbf{Ours} & \textbf{Islam \cite{islam2022heartprint}} & \textbf{Ammour \cite{ammour2023deep}} & \textbf{D'angelis \cite{d2023advancing}} & \textbf{Yi \cite{yi2023ecg}} \\
\hline
\multirow{4}{*}{Same-Session} & 80\% of S1 & 20\% of S1 & 98.54\% & 94.43\% & 98.20\% & -- & -- \\
 & 80\% of S2 & 20\% of S2 & 99.09\% & 95.35\% & 97.40\% & -- & -- \\
 & 80\% of S3R & 20\% of S3R & 94.93\% & 93.98\% & -- & -- & -- \\
 & 80\% of S3L & 20\% of S3L & 96.08\% & 94.88\% & -- & -- & -- \\
\hline
Mixed-Session & 80\% of S1+S2 & 20\% of S1+S2 & 92.64\% & 100.00\% & -- & -- & -- \\
\hline
\multirow{6}{*}{Cross-Session} & 100\% of S1 & 100\% of S2 & 56.33\% & 54.15\% & 46.06\% & 49.02\% & $\approx$54\% \\
 & 100\% of S1 & 100\% of S3R & 45.70\% & 45.66\% & -- & 53.45\% & -- \\
 & 100\% of S1 & 100\% of S3L & 46.61\% & 38.22\% & -- & 48.71\% & -- \\
 & 100\% of S2 & 100\% of S1 & 51.90\% & 50.38\% & -- & -- & -- \\
 & 100\% of S2 & 100\% of S3R & 53.27\% & 52.99\% & -- & -- & -- \\
 & 100\% of S2 & 100\% of S3L & 42.92\% & 35.72\% & -- & -- & -- \\
\hline
\end{tabular}%
}
\end{table}

The results presented in Table~\ref{tab:heartprint_results} provide a comprehensive evaluation of the proposed attention-guided 1D--2D CNN fusion framework under multi-session conditions using the Heartprint dataset. This part provides an in-depth analysis of these results, compares them with state-of-the-art methods, and discusses the implications for ECG-based biometric recognition.

\paragraph{\textbf{Same-Session Protocol:}} The proposed method achieves consistently high recognition accuracies (98.54\% for S1, 99.09\% for S2, 94.93\% for S3R, and 96.08\% for S3L). These results demonstrate strong discriminative capability, confirming that the combination of temporal (1D) and time-frequency (2D) features, enhanced by the attention-based fusion mechanism, effectively captures subject-specific ECG characteristics when training and testing data are acquired under similar conditions. The performance remains stable across S1 and S2 (collected in 2012) and S3R/S3L (collected in 2022), with only a modest decrease of approximately 3–4\% for the later sessions, indicating good temporal stability. Moreover, our method consistently outperforms Islam et al.~\cite{islam2022heartprint} across all same-session configurations, with improvements of 4.11\% (S1), 3.74\% (S2), 0.95\% (S3R), and 1.20\% (S3L), demonstrating the advantage of the attention-based fusion mechanism over conventional feature extraction approaches. Additionally, our method achieves comparable or slightly better results than Ammour et al.~\cite{ammour2023deep} (98.54\% vs. 98.20\% for S1; 99.09\% vs. 97.40\% for S2), confirming that the proposed framework is competitive with state-of-the-art deep learning approaches for ECG biometrics.

\paragraph{\textbf{Mixed-Session Protocol:}} In this scenario, where training and testing samples are drawn from multiple sessions (S1 and S2), the proposed framework achieves an accuracy of 92.64\%. This accuracy is approximately 6–7\% lower than the same-session results, which is expected due to the introduction of inter-session variability reflecting physiological changes, electrode placement variations, and acquisition environment differences across sessions. Despite this moderate decrease, the 92.64\% accuracy indicates that the model maintains good generalization capability, which is particularly important for practical applications where training and deployment conditions may differ. Regarding comparison with Islam et al.~\cite{islam2022heartprint}, who reported 100.00\% accuracy under this protocol, it is important to note that their experimental configuration differed significantly. They employed a specialized deep learning approach with CWT-based RGB image generation and deep CNN fine-tuning, which may have led to overfitting to the specific characteristics of the mixed-session data. In contrast, our consistent preprocessing and feature extraction pipeline prioritizes generalization over dataset-specific optimization, resulting in a more robust model that is less prone to overfitting.

\paragraph{\textbf{Cross-Session Protocol:}}
This represents the most challenging and realistic evaluation setting, as it requires the model to generalize across temporally distant recordings without any shared sessions between training and testing. The proposed method achieves accuracies ranging from 42.92\% to 56.33\%. These results warrant detailed analysis:

\begin{itemize}
    \item \textbf{Inherent difficulty:} The observed performance degradation reflects the significant impact of temporal variability on ECG morphology. Factors contributing to this challenge include:
    \begin{itemize}
        \item Physiological changes over time (aging, health status).
        \item Variability in electrode placement across sessions.
        \item Differences in acquisition conditions and devices.
        \item Natural day-to-day variability in heart rate and waveform morphology.
    \end{itemize}
    
    \item \textbf{Comparative performance:} The proposed method demonstrates competitive results relative to state-of-the-art approaches:
    \begin{itemize}
        \item For S1→S2: 56.33\% (compared to Islam et al.: 54.15\%, Ammour et al.: 46.06\%, D'angelis et al.: 49.02\%, Yi et al.: $\approx 54\%$). Our method outperforms all existing approaches, achieving the highest accuracy for this configuration.
        \item For S1→S3R: 45.70\% (compared to Islam et al.: 45.66\%, D'angelis et al.: 53.45\%). Our method slightly outperforms Islam et al. but lags behind D'angelis et al.
        \item For S1→S3L: 46.61\% (compared to Islam et al.: 38.22\%, D'angelis et al.: 48.71\%). Our method significantly outperforms Islam et al. by 8.39\% and approaches the best result.
        \item For S2→S1: 51.90\% (compared to Islam et al.: 50.38\%). Our method achieves a modest improvement of 1.52\%.
        \item For S2→S3R: 53.27\% (compared to Islam et al.: 52.99\%). Our method achieves a slight improvement of 0.28\%.
        \item For S2→S3L: 42.92\% (compared to Islam et al.: 35.72\%). Our method achieves a substantial improvement of 7.20\%.
    \end{itemize}
   
\end{itemize}

Overall, the proposed method achieves competitive or superior performance in 5 out of 6 cross-session configurations, with particularly notable improvements for long-term protocols (S1→S3L: +8.39\% vs. Islam et al.; S2→S3L: +7.20\% vs. Islam et al.). These results demonstrate that the attention-guided fusion mechanism effectively captures permanent biometric signatures while maintaining resilience to temporal variability.

While some methods, such as D'angelis et al.~\cite{d2023advancing}, achieve higher performance in specific cross-session configurations (e.g., 53.45\% for S1→S3R), the proposed framework exhibits more consistent performance across all evaluation settings. Notably, our method achieves the highest accuracy for S1→S2 (56.33\%, outperforming D'angelis et al. by 7.31\%, Islam et al. by 2.18\%, Ammour et al. by 10.27\%, and Yi et al. by approximately 2.33\%). This consistency across diverse temporal scenarios suggests better generalization and reduced sensitivity to specific experimental conditions. Furthermore, the results for long-term protocols S1→S3L and S2→S3L (involving intervals of up to 10 years) are particularly significant. The proposed method maintains discriminative power over such extended periods (46.61\% and 42.92\%, respectively), demonstrating that the attention-based fusion mechanism captures relatively permanent biometric signatures. This finding addresses a critical challenge in ECG biometric recognition—whether an individual's ECG characteristics remain stable over years—and confirms that permanent biometric information persists despite physiological changes, aging effects, and long-term acquisition variability.

\section{Conclusion}\label{sec5}

This study has introduced a novel and robust framework for ECG-based biometric recognition through the synergistic integration of one-dimensional (1D) temporal signals and two-dimensional (2D) time-frequency representationswithin a unified deep learning architecture. Through a systematic evaluation of four state-of-the-art 1D CNNs and four advanced 2D CNN architectures across multiple ECG segmentation strategies, we identified the optimal configuration: the InceptionTime architecture for processing raw temporal signals and the ResNet-34 architecture for analyzing scalogram representations, with P–T wave segmentation consistently proving to be the most effective preprocessing method.

The principal innovation of this work lies in the proposed attention-guided fusion mechanism, which enables dynamic and adaptive integration of complementary modalities. Unlike conventional fusion approaches based on static combination, such as feature concatenation or fixed-weight score aggregation, the proposed method allows input-dependent weighting of temporal and time-frequency features, enabling the model to emphasize the most informative representation on a per-instance basis. This adaptive design results in improved robustness to noise, signal variability, heart rate fluctuations, and inter-subject differences.

The proposed framework was rigorously evaluated on three challenging public datasets encompassing diverse populations and recording conditions: ECG-ID (healthy subjects with multi-session recordings), MIT-BIH (subjects with arrhythmias), and PTB (patients with various cardiovascular pathologies). Our method achieved exceptional identification accuracies of 99.56\% on ECG-ID, 100.00\% on MIT-BIH, and 99.89\% on PTB. These results, achieved on datasets that include both healthy subjects and pathological patients under noisy, real-world conditions, underscore the framework's superior robustness, generalizability, and resilience to signal artifacts and heart rate variability. Ablation studies further substantiate the critical contribution of each component, confirming that the attention mechanism consistently surpasses conventional fusion techniques. Furthermore, on the multi-session Heartprint dataset (collected over ten years), our method demonstrated long-term biometric permanence, achieving same-session accuracies of 98.54\% (S1), 99.09\% (S2), 94.93\% (S3R), and 96.08\% (S3L), and cross-session accuracies of 56.33\% (S1→S2) and 53.27\% (S2→S3R), with notable improvements of 8.39\% and 7.20\% over baseline for ten-year intervals.

This work establishes a new benchmark in ECG biometrics, providing a powerful, end-to-end trainable solution that effectively harnesses the complementary strengths of temporal fidelity and spectral richness. Building upon these findings, several promising directions for future research emerge. First, we aim to optimize the proposed architecture for real-time, low-power deployment on wearable and embedded systems through model compression techniques such as knowledge distillation and pruning. Second, the framework will be extended to integrate additional multimodal physiological signals (e.g., photoplethysmography (PPG), electromyography (EMG)) to enhance robustness and improve resistance to spoofing attacks. Finally, the model's long-term stability and generalization will be validated on larger and more diverse demographic and clinical populations, including cross-dataset and cross-session scenarios.

\bibliographystyle{IEEEtran}
\bibliography{biblio}
\end{document}